\pdfoutput=1

\documentclass[11pt]{article}

\usepackage[preprint]{acl}

\usepackage{times}
\usepackage{latexsym}
\usepackage{enumitem}
\usepackage{arydshln}
\usepackage{pifont}
\usepackage[T1]{fontenc}

\usepackage[utf8]{inputenc}

\usepackage{microtype}
\usepackage[normalem]{ulem}
\useunder{\uline}{\ul}{}
\usepackage{inconsolata}

\usepackage{graphicx}
\usepackage{amssymb}

\usepackage{booktabs}
\usepackage{makecell}
\usepackage{amsmath}
\usepackage{subfigure}
\usepackage{multirow}
\usepackage{multicol}
\definecolor{OliveGreen}{HTML}{3C8031}
\usepackage{adjustbox}
\usepackage{array}

\usepackage{algorithm}
\usepackage{algpseudocode}
\usepackage{colortbl}

\usepackage{amssymb}
\usepackage{mathtools}
\usepackage{amsthm}

\usepackage{colortbl}
\usepackage{nicematrix}

\usepackage{amsthm}
\usepackage{amsmath,amsfonts,bm}
\usepackage{nicefrac}
\usepackage{array}

\usepackage{url}
\usepackage{pbox}

\definecolor{bluex}{rgb}{0.27, 0.42, 0.81}
\definecolor{purplex}{HTML}{9564bf}
\definecolor{red3}{HTML}{C52A20}
\definecolor{red2}{HTML}{B36A6F}
\definecolor{red1}{HTML}{FFb5b5}
\definecolor{purple}{HTML}{B36A6F}
\definecolor{darkyellow}{HTML}{D5BA82}
\definecolor{blue1}{HTML}{508AB2}
\definecolor{blue2}{HTML}{C4E4E3}
\definecolor{green1}{HTML}{A1D0C7}
\definecolor{green2}{HTML}{BFF6BA}
\definecolor{green3}{HTML}{028100}
\definecolor{teal}{HTML}{508AB2}
\definecolor{purple1}{HTML}{8d3a94}

\usepackage{pifont}

\usepackage[listings]{tcolorbox}
\tcbuselibrary{listings,theorems}
\newtcolorbox{mybox}{colback=white!5!white,colframe=black!75!black, left=.05in, right=.05in}

\newtcbtheorem[number within=section]{exmp}{Example}%
{beforeafter skip balanced=.01in,colback=gray!5,colframe=black!75,fonttitle=\small \bfseries, left=.02in, right=.02in,bottom=.02in, top=.02in}{exmp}

\theoremstyle{plain}

\theoremstyle{definition}

\theoremstyle{remark}

\title{A Strategic Coordination Framework of Small LLMs Matches \\Large LLMs in Data Synthesis}

\author{
    Xin Gao\textsuperscript{1},
    Qizhi Pei\textsuperscript{2},
    Zinan Tang\textsuperscript{1},
    Yu Li\textsuperscript{1},
    {\bf Honglin Lin\textsuperscript{1}}, \\
    {\bf Jiang Wu\textsuperscript{1}},
    {\bf Lijun Wu\textsuperscript{1}},
    {\bf Conghui He\textsuperscript{1}\thanks{\ \ Corresponding author.}} \\
    \textsuperscript{1}Shanghai AI Laboratory,
    \textsuperscript{2}Renmin University of China \\
    \texttt{\{gaoxin,wulijun,heconghui\}@pjlab.org.cn} \\
}

\begin{document}
\maketitle
\begin{abstract}
While data synthesis and distillation are promising strategies to enhance small language models, current approaches heavily rely on Large Language Models (LLMs), which suffer from high computational costs, environmental inefficiency, and potential biases inherited from monolithic architectures. In contrast, smaller LLMs are more accessible and sustainable, but their individual capabilities often fall short in generating high-quality, diverse, and reliable data. Inspired by collaborative human processes (e.g., peer review), we propose a multiple small LLMs involved framework, \textbf{GRA}, that aggregates specialized roles across small LLMs to iterative refinement and quality control typically achieved by a single large LLM. 
In this collaborative framework, multiple small LLMs assume distinct roles—\textbf{Generator}, \textbf{Reviewer}, and \textbf{Adjudicator}—to simulate a peer-review-inspired data synthesis pipeline. The Generator proposes initial data samples, the Reviewer critiques their quality and diversity, and the Adjudicator resolves conflicts to finalize the output. By decomposing the synthesis process into specialized sub-tasks, collaborative small LLMs can achieve data-level parity with large LLM-based distillation.
Through experiments across multiple benchmarks, we demonstrate that GRA-produced data matches or exceeds the quality of single large LLM outputs, e.g., Qwen-2.5-72B-Instruct. 
Our results challenge the necessity of monolithic large models for high-quality data synthesis, advocating instead for strategic coordination of smaller agents. 
Our datasets, models, and code are publicly available at \url{https://github.com/GX-XinGao/GRA}.

\end{abstract}

\begin{figure}
    \centering
    \includegraphics[width=\linewidth]{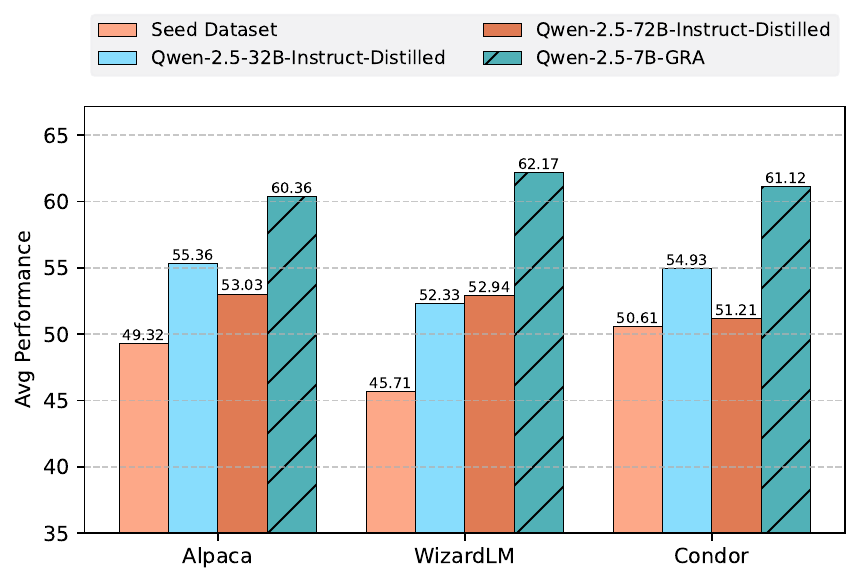}
  \caption{Average performance across GRA, vanilla seed dataset and lagrge LLMs distilled data with Qwen-2.5-7B base model.}
  \label{fig:brief_comparison}
\end{figure}

\section{Introduction}

The advent of Large Language Models (LLMs) with hundreds of billions of parameters has revolutionized artificial intelligence~\cite{llama3,qwen25,openai_gpt4,deepseek_v3}, achieving state-of-the-art (SOTA) performance across diverse domains such as natural language understanding~\cite{naveed2023comprehensive}, reasoning~\cite{deepseek_r1,openai_o1}, math problem solving~\cite{llm_math_survey}, and code generation~\cite{llm_codegen_survey}. Despite their remarkable capabilities, these monolithic models demand immense computational resources and energy, posing significant barriers to deployment in resource-constrained scenarios like edge devices or latency-sensitive applications~\cite{zheng2024review}. This paradox has spurred growing interest in developing smaller-scale LLMs that balance efficiency and effectiveness, positioning compact yet capable models as a critical research frontier in sustainable AI~\cite{chen2024role}.

To enhance small LLMs performance, knowledge distillation~\cite{hinton2015distilling}—transferring capabilities from large models via data synthesis—has emerged as a dominant paradigm~\cite{xu2024survey}. However, conventional distillation inherently depends on the availability of powerful LLMs~\cite{wang2024survey}, which introduces practical limitations: deploying large LLMs for data generation incurs prohibitive costs, especially in privacy-sensitive or security-critical scenarios requiring local execution~\cite{hagos2024recent,yao2024survey}. While small LLMs offer computational accessibility, their individual capacities are inherently constrained, struggling to match the quality, diversity, and reliability of data synthesized by their larger counterparts. This raises a pivotal question: \textit{Can we design a resource-efficient framework that enables small models to collectively rival the distillation capabilities of monolithic LLMs?}

To address this challenge, we propose \textbf{GRA}, a collaborative framework inspired by human peer-review mechanisms. Diverging from traditional single-model distillation, GRA orchestrates multiple specialized small LLMs into three synergistic roles: (1) \textbf{Generator}, which proposes candidate data samples; (2) \textbf{Reviewer}, which evaluates quality and diversity through iterative critiques; and (3) \textbf{Adjudicator}, which resolves conflicts to finalize outputs. 
Notably, the synthetic data generated by a single Generator will be evaluated by multiple Reviewers, and the Adjudicator is to resolve significant disagreements among the Reviewers by reassessing the data.
By decomposing the synthesis pipeline into discrete sub-tasks and enabling role-specific optimization, GRA mimics the refinement processes typically concentrated within a single large LLM. This collaborative architecture not only circumvents the resource-intensity of large models but also amplifies small models’ collective intelligence through task specialization and iterative quality control—effectively operationalizing the “wisdom of crowds” principle in model ensembles.

We validate GRA through extensive experiments on multiple domains, including general question answering, reasoning, mathematical and coding tasks. 
Remarkably, our framework achieves data quality comparable to or exceeding outputs from SOTA monolithic LLMs like Qwen-2.5-72B-Instruct, while requiring orders-of-magnitude fewer computational resources. These results empirically demonstrate that strategically coordinated small LLMs can collectively surpass the distillation performance of individual large LLMs.

\section{Related Work}

\subsection{Data Synthesis}

The rapid evolution of LLMs has increased the demand for high-quality, diverse training data, creating a bottleneck in AI development~\cite{zhao2023survey}. Synthetic data is a promising direction to improve data. 
A dominant paradigm for synthetic data leverages knowledge distillation from LLMs. WizardLM~\cite{xu2023wizardlm} iteratively enhances seed data based on  predefined evolutionary rules and instructions.
MAGPIE~\cite{magpie} utilizes pre-defined query templates to directly prompt the LLM to synthesize data while not relying on seed data. MuggleMath~\cite{muggle_math} employs LLMs to rewrite and enhance seed datasets by generating new variations of problem statements and corresponding answers.

In addition to single-LLM-based approaches, several works also employ multiple agents for data generation~\cite{agentinstruct,geneticinstruct}. AgentInstruct~\cite{agentinstruct} leverages GPT-4~\cite{openai_gpt4} to facilitate data generation and curation through the use of multiple agents. Similarly, Genetic Instruct~\cite{geneticinstruct} defines three distinct roles—Instructor-LLM, Coder-LLM, and Judge-LLM—to generate code data. Unlike previous multi-agent frameworks that rely on a single LLM like GPT-4 for repeated deployment across different stages, GRA pioneers a collaborative paradigm that coordinates diverse types of small LLMs through specialized role allocation to collectively rival the synthetic data quality of singular large LLMs.

\subsection{Data Curation} 
The efficacy of synthetic data critically hinges on post-synthesis curation~\cite{long2024llms}. Early curation methodologies predominantly relied on surface-level heuristics, such as repetition counts~\cite{self_repetition} and lexical diversity metrics~\cite{ngram,texygen,shu2019generating}, which inadequately captured semantic coherence and task-specific utility.
Recent works shift toward LLM-driven curation frameworks. Initial efforts employed perplexity (PPL) and instruction following difficulty score(IFD)~\cite{ifd} to filter low-quality samples. The emergence of ``LLM-as-judge'' paradigms further revolutionized this domain, leveraging well-tuned LLMs for quality assessment~\cite{gu2024survey}. In contrast,  while sharing the ``LLM-as-judge'' basis,  our GRA framework differs by using multiple specialized small LLMs for diverse evaluations. Unlike single-LLM approaches that risk bias, GRA employs a multi-agent system and a two-tier adjudication to ensure balanced and comprehensive data assessment.

\section{Methodology}
\begin{figure*}[!htbp]
  \centering
  \includegraphics[width=\linewidth]{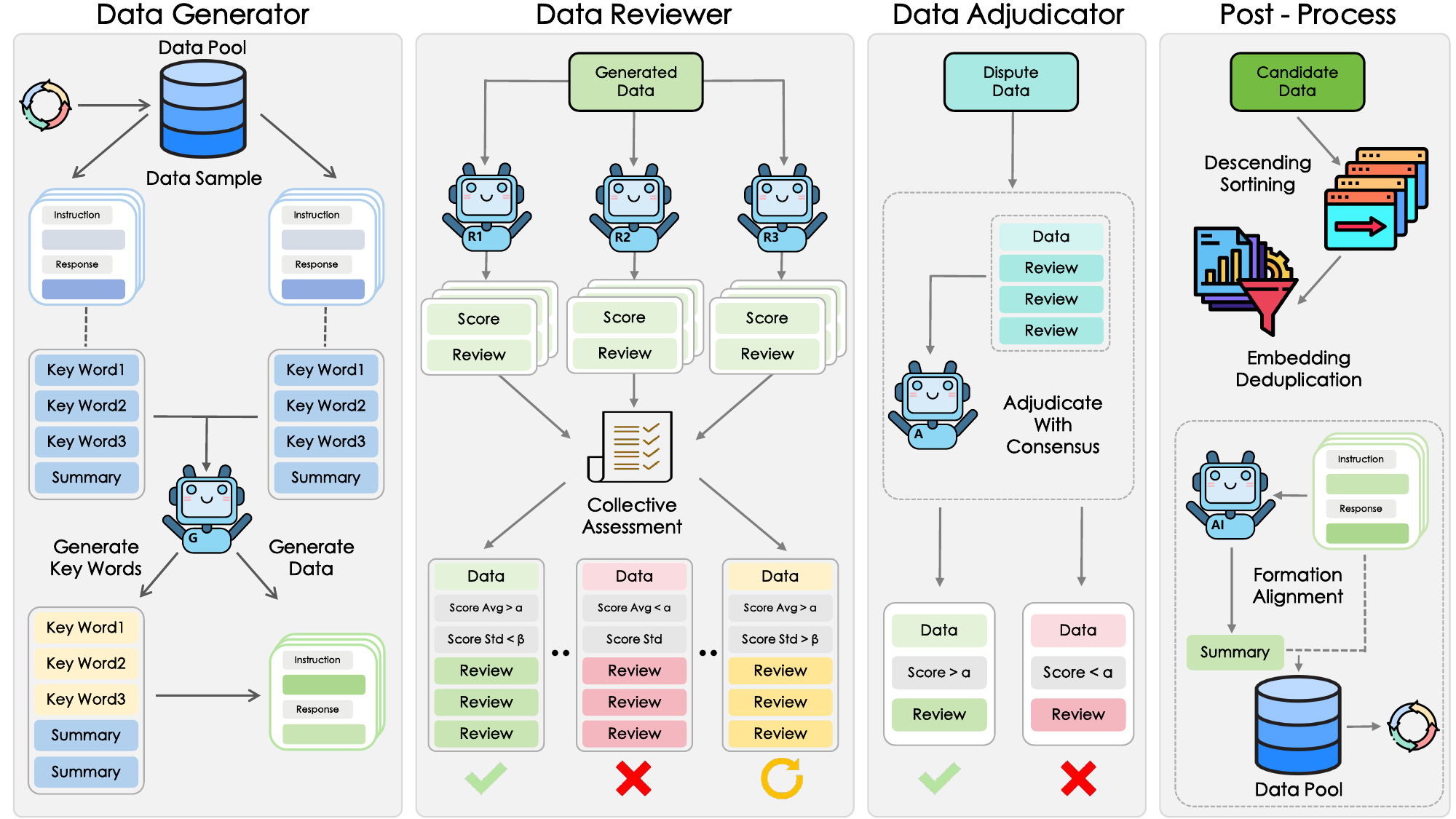}
  \caption{Overview of GRA's architecture, highlighting its four key modules: (a) The Generator creates domain-specific samples, (b) followed by collaborative evaluation by Reviewers, (c) The Adjudicator resolves conflicts, and (d) Post-Processing refines the results by removing redundancies.}
  \label{fig:overview}
\end{figure*}

\paragraph{Preliminary}
We first provide essential background and definitions for key components with the rationale behind our design choices.
Specifically, we denote \( \mathcal{M} \) as the pool of small LLMs employed in the framework, and data pool at iteration \( t \) as \( \mathcal{D}^{(t)} = \{(x_i, y_i, d_i, k_i, s_i)\}_{i=1}^N \),
where each sample in \(\mathcal{D}^{(t)} \) consists of an instruction \( x_i \), a response \( y_i \), a domain label \( d_i \), a 3-tuple of keywords \( k_i \), and a summary \( s_i \).
The domain label \( d_i \) indicates the task type of the sample, which is categorized into seven domains: Coding, Math, Question and Answering, Reasoning, Role Play, Language, and Creation, which are designed to aim for covering the most general domains. 
The 3-tuple keywords \( k_i \) are generated by LLMs to capture the core concepts of each data sample.
In addition, we include a summary \( s_i \) to provide context and clarify keyword relationships, enhancing the overall understanding of the data.
More details such as domain definitions and construction of \(\mathcal{D}^{(0)} \)-seed dataset are shown in the Appendix~\ref{sec:app_gen}.

\subsection{Pipeline Overview}
The overview of our GRA is shown in Figure~\ref{fig:overview}. 
GRA integrates domain-specific generation with collaborative curation by orchestrating three specialized roles. First, the Generator produces initial samples using few-shot prompting tailored to the domain. Next, multiple Reviewers collaboratively assess the quality of these samples. Finally, an Adjudicator resolves any conflicting evaluations through consensus-driven arbitration. 
A post-processing module further refines the results by eliminating redundant outputs via embedding deduplication and metadata enrichment. 
Moreover, to prevent model specialization bias, GRA dynamically assigns distinct LLMs from $\mathcal{M}$ to each role through random selection and during generation.

\subsection{Data Generator}  

To enhance synthetic data diversity while maintaining domain consistency, we carefully design the generation process to systematically integrate conceptual knowledge and contextual patterns.  
For each instance, we randomly select one LLM $M_{G} \in \mathcal{M}$ as the Generator. For a given domain, the generation workflow comprises three coordinated stages: 
First, $M_{G}$ analyzes randomly selected keywords-summary pairs ($k_i, s_i$) from the current domain corpus, synthesizing a novel 3-tuple keywords $k'$ by extracting and combining meaningful semantic patterns and attributes from the input data. Second, drawing inspiration from contextual summaries of seed examples, $M_{G}$ contextualizes the synthesized keywords into an executable instruction $x'$. Last, $M_{G}$ produces response $y'$ that appropriately addresses the generated instruction $x'$.

\subsection{Data Reviewer}  
To ensure equitable data evaluation and mitigate individual model biases, the Reviewer module performs a rigorous multi-agent quality assessment through coordinated evaluation protocols. 
We randomly select $N_R$ reviewers from $\mathcal{M} \setminus {M_G}$ to form the review committee $R=\{M_{R_i}\}_{i=1}^{N_R}$.

The review process includes two stages: \textit{instruction evaluation} and \textit{response evaluation}.
During the \textit{instruction evaluation} phase, each Reviewer independently verifies instruction $x'$ against three binary criteria: ``Reasonableness, Completeness, and Clarity''.
Immediate rejection occurs if any Reviewer $M_{R_i}$ flags any criterion failure.
Approved instructions then advance to the \textit{response evaluation} stage, where each Reviewer conducts a granular analysis of the generated response $y'$ from six dimensional metrics: ``Correctness, Clarity, Completeness, Relevance, Coherence, and Ethicality''. 
Then each Reviewer quantitatively assesses these dimensions by assigning integer scores from 0 to 10 and the averaged score serves as the final score.
Additionally, each Reviewer provides detailed textual feedback that delineates both the strengths and weaknesses of the generated response.

In the end, the committee $R$ aggregates assessments through calculating the mean composite score $\mu_R$ and assessing consensus reliability through standard deviation $\sigma_R$. After all, the committee $R$ makes the final decision: (a) \textbf{Rejection}: If \( \mu_R < \tau \). (b) \textbf{Acceptance}: If \( \mu_R \geq \tau \) with \( \sigma_R \leq \delta\). (c). \textbf{Adjudication}: If \( \mu_R \geq \tau \) but \( \sigma_R > \delta\), where $\tau $ and $\delta$ are predefined thresholds.

\subsection{Data Adjudicator}  

Adjudicator is designed for reassessing the disputed samples. 
For each disputed sample, we randomly select an Adjudicator $M_A$ from $\mathcal{M} \setminus ({M_G} \cup R)$ to perform consensus-driven arbitration. For each disputed sample, based on consolidating Reviewers' feedback, $M_A$ conducts a comprehensive reassessment using the same evaluation criteria as the Reviewers and gets the final composite score $s_A$.
Only samples with $s_A \geq \tau$ proceed to the subsequent post-processing module.

\subsection{Post-Processing Module}  
 
The post-processing module ensures syntactic consistency and enhances diversity for final dataset integration. 
First, we generate embeddings for all data using the all-MiniLM-L6-v2 encoder~\cite{reimers-2019-sentence-bert}. 
Data are processed based on their review score $\mu_R$ in descending order. Each sample, along with its embedding, undergoes semantic redundancy verification against the target data pool: if the maximum cosine similarity with any item in the pool is below a specified similarity threshold, the candidate is added to the pool. Otherwise, it is rejected to prevent semantic duplication. The deduplicated data then undergoes metadata enrichment through random LLM $M_P \in \mathcal{M}$ to generate contextual summary $s'$, supplemented with the keywords from which it was derived, ensuring format consistency with the existing data. This two-stage filtration—embedding-based semantic deduplication followed by LLM summarization—ensures both conceptual diversity and structural compatibility for subsequent data synthesis workflows.

\section{Experiment}

\subsection{Settings}

\paragraph{GRA Data Generation.}
Our framework operates on three established instruction-tuning datasets: Alpaca~\cite{alpaca}, WizardLM~\cite{xu2023wizardlm}, and the recent Condor~\cite{cao2025condor}, which serve as seed datasets for data synthesis. As for the small LLMs, we use five diverse and representative LLMs within 7B-8B parameters: Llama-3.1-8B-Instruct~\cite{dubey2024llama}, Qwen2.5-7B-Instruct~\cite{yang2024qwen2}, InternLM3-8B-Instruct~\cite{cai2024internlm2}, Mistral-7B-Instruct-v0.3~\cite{jiang2023mistral}, and Tulu-3-8B~\cite{lambert2024t}. The synthesis process runs for five iterations, and each iteration produces approx $10$K high-quality samples per round ($51$K in total) using vLLM~\cite{kwon2023efficient} toolkit. We set the minimum score threshold $\tau = 8$ and the maximum standard deviation $\delta = 1.5$ at review stage. For deduplication, the maximum cosine similarity is set to $0.9$.
More detailed generation settings are shown in Appendix~\ref{sec:app_gen}.

\paragraph{Training.}

After data synthesis, we conduct full parameter Supervised Fine-Tuning (SFT) on two well-known base models for verification: Llama-3.1-8B-Base~\cite{dubey2024llama} and Qwen2.5-7B-Base~\cite{yang2024qwen2}. The experiments utilize the Llama-Factory~\cite{zheng2024llamafactory} toolkit. We run five iterations, each accumulating synthetic data from Rnd$1$ to Rnd$5$, representing data from the first round up to all five rounds. In the main experiment, we report the iteration with the best average performance. Detailed training settings are provided in Appendix~\ref{sec:app_train}.

\paragraph{Evaluation.}
\label{sec:evaluation}

For evaluation, following~\cite{xu2023wizardlm,cao2025condor}, we select multiple benchmarks on various domains-mathematics, coding, reasoning, and general-to demonstrate the effectiveness of our GRA framework, which include GSM8K, MATH, HumanEval, MBPP, HellaSwag, ARC-C, GPQA, IFEval, BBH and MMLU. We use the opencompass~\cite{2023opencompass} toolkit for a fair evaluation. More detailed evaluation settings are shown in Appendix~\ref{sec:app_eval}.

\begin{table*}[htbp]
\renewcommand{\arraystretch}{1.15}
\resizebox{\textwidth}{!}{
\begin{tabular}{llccccccccccccccc}
\hline
\multicolumn{1}{c}{}            &  & \multicolumn{2}{c}{\textbf{Math}} &  & \multicolumn{2}{c}{\textbf{Coding}} &                 & \multicolumn{4}{c}{\textbf{Reasoning}}     &                 & \multicolumn{2}{c}{\textbf{General}} &                 &       \\ \cline{3-4} \cline{6-7} \cline{9-12} \cline{14-15}
\textbf{Dataset }                        &  & GSM8K       & MATH       &  & Humaneval      & MBPP      &                 & HellaSwag & ARC-C & GPQA  & BBH   &                 & MMLU         & IFEval       &                 & \textbf{AVG }  \\ \hline
\multicolumn{17}{c}{\textbf{\textit{ \ding{72} Qwen-2.5-7B Base Model}}}                                                                                                                                 \\ \midrule
Alpaca                          &  & 60.12       & 27.00      &  & 32.32          & 61.20     &                 & 72.51     & 65.76 & 23.74 & 60.42 &                 & 45.19        & 44.95        &                 & 49.32 \\
\textbf{Qwen-2.5-7B-GRA }                &  & \cellcolor[HTML]{d2e6fa}70.28       & 33.82      &  & \cellcolor[HTML]{d2e6fa}76.83          & 64.80     &                 & \cellcolor[HTML]{d2e6fa}80.23     & \cellcolor[HTML]{d2e6fa}74.24 & \cellcolor[HTML]{d2e6fa}24.24 & \cellcolor[HTML]{d2e6fa}60.57 &                 & \cellcolor[HTML]{d2e6fa}65.42        & 53.16        &                 & \cellcolor[HTML]{d2e6fa}\textbf{60.36} \\
Qwen-2.5-32B-Instruct-Distilled &  & 63.23       & \cellcolor[HTML]{d2e6fa}56.26      &  & 76.22          & \cellcolor[HTML]{d2e6fa}65.40     &                 & 72.02     & 56.61 & 20.71 & 42.04 &                 & 49.05        & \cellcolor[HTML]{d2e6fa}53.71        &                 & 55.36 \\
Qwen-2.5-72B-Instruct-Distilled &  & 50.64       & 55.80      &  & 74.39          & 64.40     &                 & 74.70     & 56.95 & 23.74 & 38.75 &                 & 37.37        & 53.33        &                 & 53.03 \\ \hline
WizardLM                        &  & 64.37       & 31.94      &  & 50.61          & 64.20     &                 & 65.89     & 40.68 & 21.72 & 39.34 &                 & 31.18        & 47.12        &                 & 45.71 \\
\textbf{Qwen-2.5-7B-GRA }                &  & 58.98       & 47.84      &  & 76.83          & 64.60     &                 & \cellcolor[HTML]{d2e6fa}81.58     & \cellcolor[HTML]{d2e6fa}79.66 & \cellcolor[HTML]{d2e6fa}28.79 & \cellcolor[HTML]{d2e6fa}63.99 &                 & \cellcolor[HTML]{d2e6fa}65.23        & \cellcolor[HTML]{d2e6fa}54.15        &                 & \cellcolor[HTML]{d2e6fa}\textbf{62.17} \\
Qwen-2.5-32B-Instruct-Distilled &  & 59.36       & 54.96      &  & 72.56          & \cellcolor[HTML]{d2e6fa}66.20     &                 & 54.22     & 54.58 & 21.21 & 37.63 &                 & 46.99        & 53.96        &                 & 52.33 \\
Qwen-2.5-72B-Instruct-Distilled &  & \cellcolor[HTML]{d2e6fa}69.83       & \cellcolor[HTML]{d2e6fa}56.30      &  & \cellcolor[HTML]{d2e6fa}79.88          & 65.20     &                 & 57.85     & 56.61 & 16.67 & 37.02 &                 & 35.02        & 53.96        &                 & 52.93 \\ \hline
Condor                          &  & 57.16       & 48.60      &  & 51.83          & 64.60     &                 & 55.74     & 66.10 & 24.75 & 39.96 &                 & 47.48        & 49.86        &                 & 50.61 \\
\textbf{Qwen-2.5-7B-GRA  }               &  & 64.75       & 42.82      &  & \cellcolor[HTML]{d2e6fa}76.22          & 65.60     &                 & \cellcolor[HTML]{d2e6fa}79.90     & \cellcolor[HTML]{d2e6fa}75.59 & \cellcolor[HTML]{d2e6fa}26.26 & \cellcolor[HTML]{d2e6fa}61.49 &                 & \cellcolor[HTML]{d2e6fa}64.59        & \cellcolor[HTML]{d2e6fa}54.02        &                 & \cellcolor[HTML]{d2e6fa}\textbf{61.12} \\
Qwen-2.5-32B-Instruct-Distilled &  & \cellcolor[HTML]{d2e6fa}71.34       & \cellcolor[HTML]{d2e6fa}56.00      &  & 68.90          & \cellcolor[HTML]{d2e6fa}66.40     &                 & 63.93     & 61.02 & 18.69 & 37.53 &                 & 54.56        & 52.95        &                 & 54.93 \\
Qwen-2.5-72B-Instruct-Distilled &  & 62.22       & 54.46      &  & 75.61          & 65.40     &                 & 61.99     & 52.88 & 14.14 & 32.18 &                 & 39.61        & 52.41        &                 & 51.21 \\  \midrule

\multicolumn{17}{c}{\textbf{\textit{\ding{115} Llama-3.1-8B Base Model}}}                                                                                                                                                                         \\ \hline
Alpaca                          &  & 51.71       & 11.32      &  & 43.29          & 45.20     &                 & 53.25     & \cellcolor[HTML]{e2f0da}75.25 & 21.21 & 61.86 &                 & \cellcolor[HTML]{e2f0da}52.65        & 43.06        &                 & 45.88 \\
\textbf{Llama-3.1-8B-GRA }               &  & 65.58       & 17.20      &  & \cellcolor[HTML]{e2f0da}52.44          & 48.60     &                 & 66.30     & 71.53 & \cellcolor[HTML]{e2f0da}29.80  & 64.50  &                 & 48.58        & 56.80         &                 & 52.13 \\
Qwen-2.5-32B-Instruct Distilled &  & 64.44       & 23.32      &  & 48.17          & 49.20     &                 & \cellcolor[HTML]{e2f0da}69.14     & 69.49 & 26.77 & \cellcolor[HTML]{e2f0da}65.07 &                 & 45.78        & 55.73        &                 & 51.71 \\
Qwen-2.5-72B-Instruct Distilled &  & \cellcolor[HTML]{e2f0da}67.40       & \cellcolor[HTML]{e2f0da}23.82      &  & 48.78          & \cellcolor[HTML]{e2f0da}50.60     &                 & 63.95     & 74.24 & 26.77 & 64.77 &                 & 48.09        & \cellcolor[HTML]{e2f0da}58.39        &                 & \cellcolor[HTML]{e2f0da}\textbf{52.68} \\ \hline
WizardLM                        &  & 63.84       & 16.78      &  & 45.73          & 47.00     &                 & 67.16     & 63.05 & 9.60   & 62.98 &                 & 40.89        & 48.90        &                 & 46.59 \\
\textbf{Llama-3.1-8B-GRA }               &  & 65.88       & 18.98      &  & 51.22          & 47.00     &                 & 66.54     & \cellcolor[HTML]{e2f0da}72.88 & \cellcolor[HTML]{e2f0da}26.77 & 63.28 &                 & 43.10        & 53.52        &                 & 50.92 \\
Qwen-2.5-32B-Instruct Distilled &  & 64.97       & 23.50      &  & \cellcolor[HTML]{e2f0da}51.83          & 48.00     &                 & \cellcolor[HTML]{e2f0da}67.66     & 65.08 & 23.74 & \cellcolor[HTML]{e2f0da}65.03 &                 & 43.31        & 54.51        &                 & 51.10 \\
Qwen-2.5-72B-Instruct Distilled &  & \cellcolor[HTML]{e2f0da}66.64       & \cellcolor[HTML]{e2f0da}24.12      &  & \cellcolor[HTML]{e2f0da}51.83          & \cellcolor[HTML]{e2f0da}48.60     &                 & 63.76     & 69.15 & \cellcolor[HTML]{e2f0da}26.77 & 63.60 &                 & \cellcolor[HTML]{e2f0da}46.98        & \cellcolor[HTML]{e2f0da}56.28        &                 & \cellcolor[HTML]{e2f0da}\textbf{52.18 }\\ \hline
Condor                          &  & 61.49       & 22.12      &  & 38.41          & 48.40     &                 & 47.24     & 51.53 & 15.66 & 65.28 &                 & 27.91        & 52.35        &                 & 43.04 \\
\textbf{Llama-3.1-8B-GRA }               &  & 65.58       & 15.82      &  & 50.61          & 48.80     &                 & 69.22     & \cellcolor[HTML]{e2f0da}72.54 & \cellcolor[HTML]{e2f0da}26.77 & 63.48 &                 & \cellcolor[HTML]{e2f0da}45.22        & 53.54        &                 & 51.60 \\
Qwen-2.5-32B-Instruct Distilled &  & 64.94       & 23.00      &  & \cellcolor[HTML]{e2f0da}54.88          & 49.80     &                 & \cellcolor[HTML]{e2f0da}71.04     & 65.08 & 24.24 & \cellcolor[HTML]{e2f0da}66.12 &                 & 44.37        & 52.64        &                 & \cellcolor[HTML]{e2f0da}\textbf{51.81} \\
Qwen-2.5-72B-Instruct Distilled &  & \cellcolor[HTML]{e2f0da}69.14       & \cellcolor[HTML]{e2f0da}23.30      &  & 53.05          & \cellcolor[HTML]{e2f0da}50.20     &                 & 67.40     & 68.14 & 21.72 & 63.71 &                 & 45.02        &\cellcolor[HTML]{e2f0da}53.94        &                 &51.56\\  \midrule
 \bottomrule
\end{tabular}
}
\caption{Main Comparison between our proposed method, the vanilla seed datasets, and the data distilled from large-scale LLMs (Qwen-2.5-32B-Instruct and Qwen-2.5-72B-Instruct). Colored values represent the best results.}
\label{tab:main_result}
\end{table*}

\subsection{Main Results}
The main results are presented in Table~\ref{tab:main_result}, where we compare GRA against vanilla seed datasets and also single large LLM (Qwen2.5-32B/72B-Instruct) distillation.
From the table, we can draw following observations:
\paragraph{Superiority Over Vanilla Seed Data.} The GRA method achieves significant improvements over vanilla seed data, with average performance gains of $6.18$\% (Llama-3.1-8B) and $11.81$\% (Qwen-2.5-7B) across all benchmarks.
These results demonstrate the effectiveness of GRA.
We attribute this success to GRA's ability to enhance data diversity and quality through iterative data synthesis. Further analysis is provided in Section~\ref{sec:analysis_quality}.
\paragraph{Competitive Against Large LLM Distillation.} GRA's collaborative small LLM framework demonstrates parity with or superiority to single large LLM distillation. Notably, comparing to Qwen-2.5-72B-Instruct distilled data, it narrows the performance gap to $0.59$\% for Llama-3.1-8B while surpassing by $8.83$\% for Qwen-2.5-7B on average. This validates the viability of cost-effective multi-agent synthesis as a scalable alternative to resource-intensive large-model distillation.
\paragraph{Base Model Dependency.} Performance varies substantially with base model capability: Qwen-based implementations consistently outperform Llama counterparts by ~9\% under identical data conditions. We attribute this discrepancy to Qwen’s strong base model ability, as demonstrated by many recent models~\cite{yang2024qwen2}, which likely enhance its ability to leverage synthesized data.
\paragraph{Diminishing Returns in Teacher Scaling.} Despite doubling parameter counts (Qwen-2.5-32B-Instruct$\to$Qwen-2.5-72B-Instruct), teacher model scaling yields marginal quality gains. 
This suggests a potential saturation point in knowledge transfer efficiency for standard distillation methods, emphasizing the need for more advanced distillation protocols rather than naive distillation from larger models.
These insights collectively highlight GRA’s practical value for efficient, high-quality data generation while delineating fundamental boundaries in LLM knowledge transfer.

\subsection{Ablation}

\begin{figure}[!t]
    \centering
    \includegraphics[width=\linewidth]{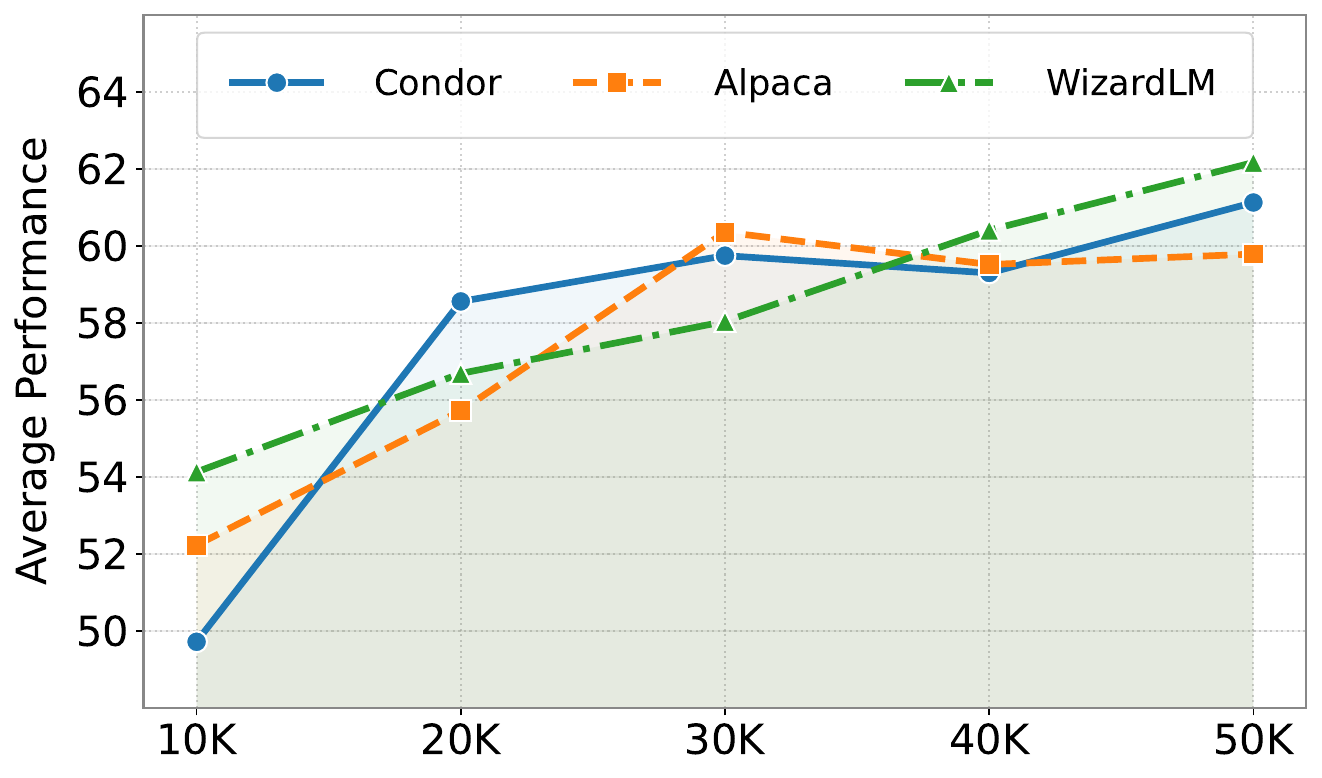}
    \caption{Performance along data iterations with Qwen-2.5-7B-Base model.}
    \label{fig:data-iteration}
\end{figure}

\paragraph{Performance along Data Iteration.}
To investigate the impact of iterative data synthesis rounds on model performance, we conduct five sequential iterations of data generation using GRA, with approx $10$K synthetic samples produced per iteration. We trained models incrementally on cumulative datasets: $10$K ($1$st iteration), $20$K (1st–2nd), $30$K ($1$st–$3$rd), $40$K ($1$st–4th), and $50$K ($1$st–$5$th). Results are illustrated in Figure~\ref{fig:data-iteration}. Experiments across three distinct seed datasets consistently demonstrate a positive correlation between iteration count and model performance. Besides, we observe that performance gains are rapid in early iterations, but gradually plateaued in later iterations, suggesting diminishing yet meaningful returns. This trend aligns with the hypothesis that initial cycles resolve critical data scarcity, whereas subsequent rounds exhaust exploitable synthetic data diversity.

\begin{figure}[!t]
    \centering
    \includegraphics[width=0.9\linewidth]{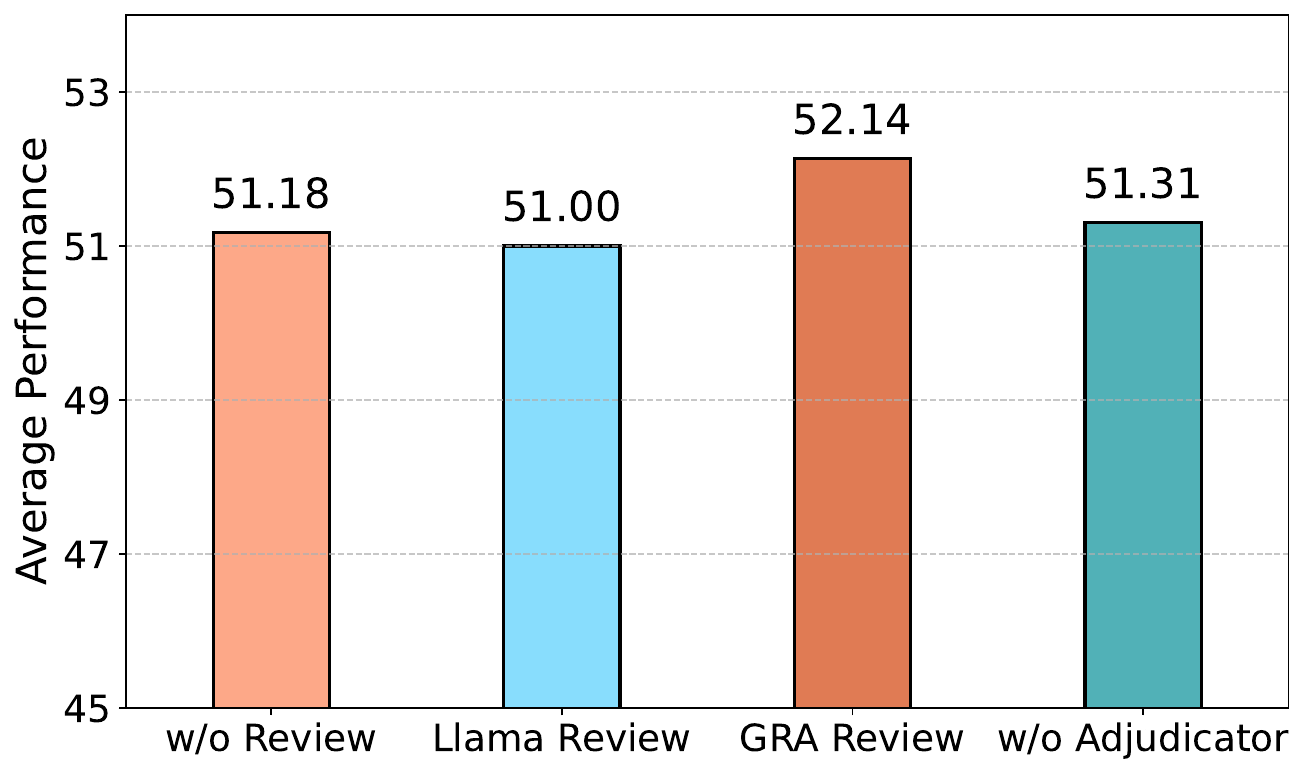}
    \caption{comparison across different setting of reviewer and adjudicator, with alpaca as seed dataset and Llama-3.1-8B as base model.}
    \label{fig:RA}
\end{figure}

\paragraph{Effect of GRA Data Reviewer.}
We conduct ablation experiments to evaluate the effectiveness of GRA’s multi-model review mechanism by comparing against three baselines: no Reviewer, Llama3.1-8B-Instruct Reviewer. As shown in Figure~\ref{fig:RA}, GRA outperforms all baselines, confirming the necessity of collaborative curation. Notably, single-model reviews show negligible gains over no review, implying either (a) limited quality-assessment capacity in small LLMs or (b) data quality from GRA’s generation phase being sufficiently robust for standalone use. These results validate that GRA’s multi-agent framework uniquely synthesizes complementary curation strengths, avoiding the pitfalls of isolated or capacity-constrained reviewers.
Comparisons wiht large LLM reviwer are shown in Section~\ref{sec:analysis_quality}.

\vspace{-0.5cm}
\paragraph{Effect of GRA Data Adjudicator.}
To validate the necessity of the Adjudicator module in GRA, we conducted ablation experiments without the Adjudicator, where conflicting reviews were resolved via majority voting. Using Llama-3.1-8B as the base model and Alpaca as the seed dataset, the Adjudicator-enhanced version achieved a 0.73\% improvement in synthetic data quality (measured by human-aligned correctness scores), demonstrating its effectiveness, as shown in Figure~\ref{fig:RA}.

\section{Analysis}

\subsection{Effect of Small LLMs' Role}
We are interested in the specific role played in the GRA framework, therefore we conduct experiments of three role assignment strategies for GRA pipeline with Qwen-2.5-7B-Base as the training model: (1) Single-Model: all roles assigned to a single LLM (Llama3.1-8B-Instruct); (2) Fixed-assignment strategy: five distinct LLMs assigned to roles in fixed permutations; (3) Our GRA setting with random roles. 
The results are as follows: when we simply use one model to play all roles (Setting 1), the average accuracy is 52.07, while Setting 2, which assigns fixed roles to five dedicated models, results in a lower accuracy of 48.59. Lastly, our GRA based training achieves 59.52 accuracy. Lastly, our GRA based training achieves 59.52 accuracy. 
From the above results, we can find that the advantages of our randomized role assignment in our GRA is clear and significant, compared to both other settings. Besides, interestingly, fixed assignments of different models performs worse than single role model, which means that self-evolve is better than fixed role assignment among different models, further demonstrating that multi-agent corporation is curial.

\subsection{Inconsistent Data Analysis}

\setcounter{table}{0}
\renewcommand{\tablename}{Case}
\begin{table}
\begin{tcolorbox}[colback=gray!5,colframe=black!75, title=GRA Data Synthesis Cases]
\small
\# Data: \\
\#\# Instruction: \\
Given a set of numbers from 1 to 100, find the sum of all numbers that are divisible by both 3 and 5. Express the sum in LaTeX format.\\
\#\# Response: \\
The sum of all numbers from 1 to 100 that are divisible by both 3 and 5 can be found by calculating the sum of the arithmetic series where the first term is $(3 \times 5 = 15)$ and the last term is $(100)$... the sum of all numbers from 1 to 100 that are divisible by both 3 and 5 is 90. In LaTeX format, this is written as: $[sum_{i=1}^{100} (3 \times 5)^i = 90]$\\

\# Data Review\\
\#\#Reviewer 1\\
detail scores : [9,10,10,10,10,10]. review comment: NO MISSTATEMENT, ALL INFO PRESENT.\\
\#\#Reviewer 2\\
detail scores : [9,9,10,10,10,10]. review comment: Accurate and well-structured response, clear and correct explanation, irrelevant information avoided, perfect coherence and suitable LaTeX expression\\
\#\#Review 3\\
detail scores : [6,4,5,4,5,3]. review comment: Incorrect arithmetic series calculation and LaTeX format error\\
\#\#Review Result\\
average score of 8 meets the threshold. score std of 2.4758 exceeds the threshold.\\

\#Data Adjudicator\\
detail scores : [4,2,5,5,5,1,5]. average score of 4.5 below threshold.\\
\#\#Adjudication Result\\
Discard unqualified data.
\end{tcolorbox}
\vspace{-0.4cm}
\caption{A case analysis of inconsistent data.}
\vspace{-0.6cm}
\label{inconsistent data}
\end{table}

We show a Case~\ref{inconsistent data} to further illustrate the effectiveness of Adjudicator. 
\setcounter{table}{1}
\renewcommand{\tablename}{Table}
For a wrong math problem, two of three reviewers incorrectly assigned high scores, while one correctly flagged the error. Although the average score ($8/10$) initially met the retention threshold, the high score variance ($2.48$) triggered adjudication. The Adjudicator synthesized all reviews and evaluation criteria, overriding the majority’s misjudgment to assign a low score ($4.5/10$), thereby filtering out the flawed data. This highlights how the Adjudicator mitigates limitations of simple voting or averaging by dynamically reconciling disagreements and enforcing rigorous quality standards.

\begin{figure}[!t]
    \centering
    \begin{minipage}{\linewidth}
        \centering
        \includegraphics[width=0.8\linewidth]{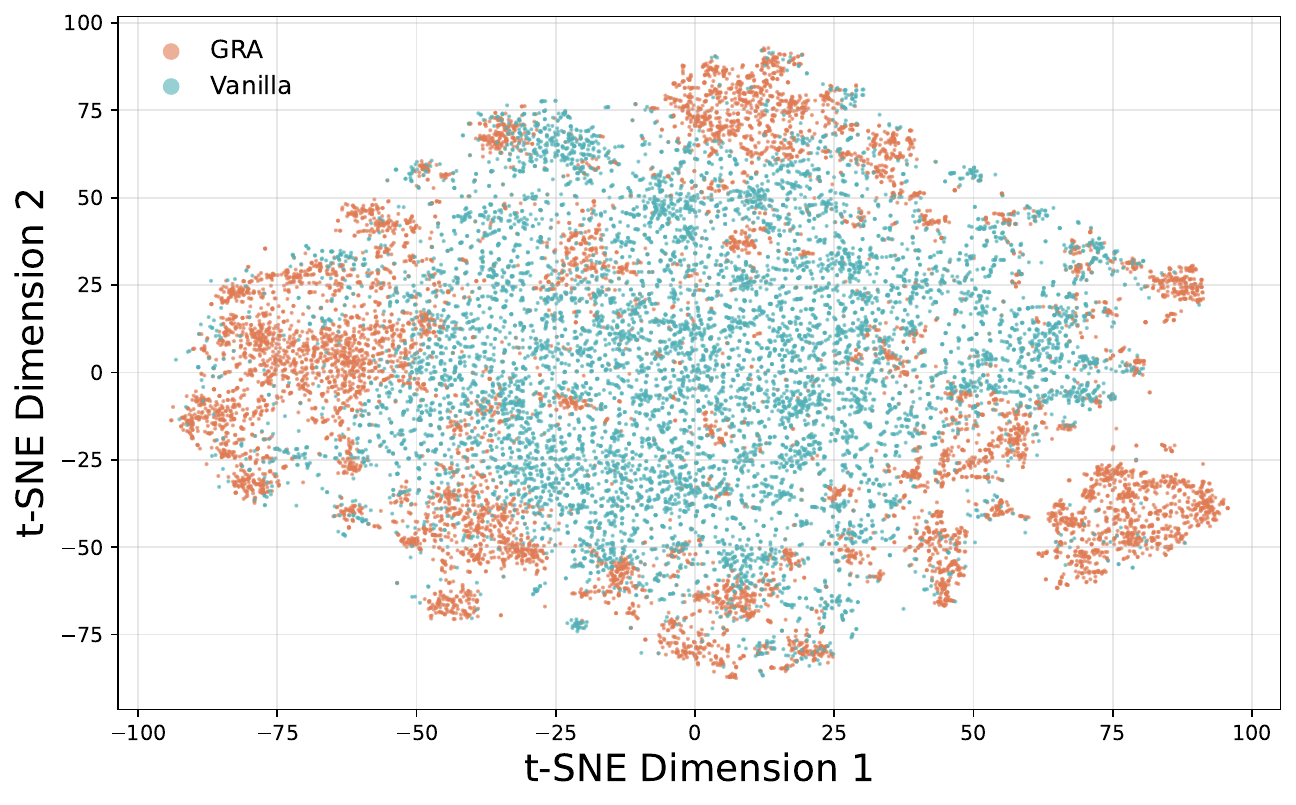}
        \vspace{-0.4cm}
        \caption{Data coverage comparison between vanilla seed dataset and GRA synthetic data.}
        \label{fig:tsne_seed_GRA}
    \end{minipage}
    \vspace{0.5cm} 
    \begin{minipage}{\linewidth}
        \centering
        \includegraphics[width=0.8\linewidth]{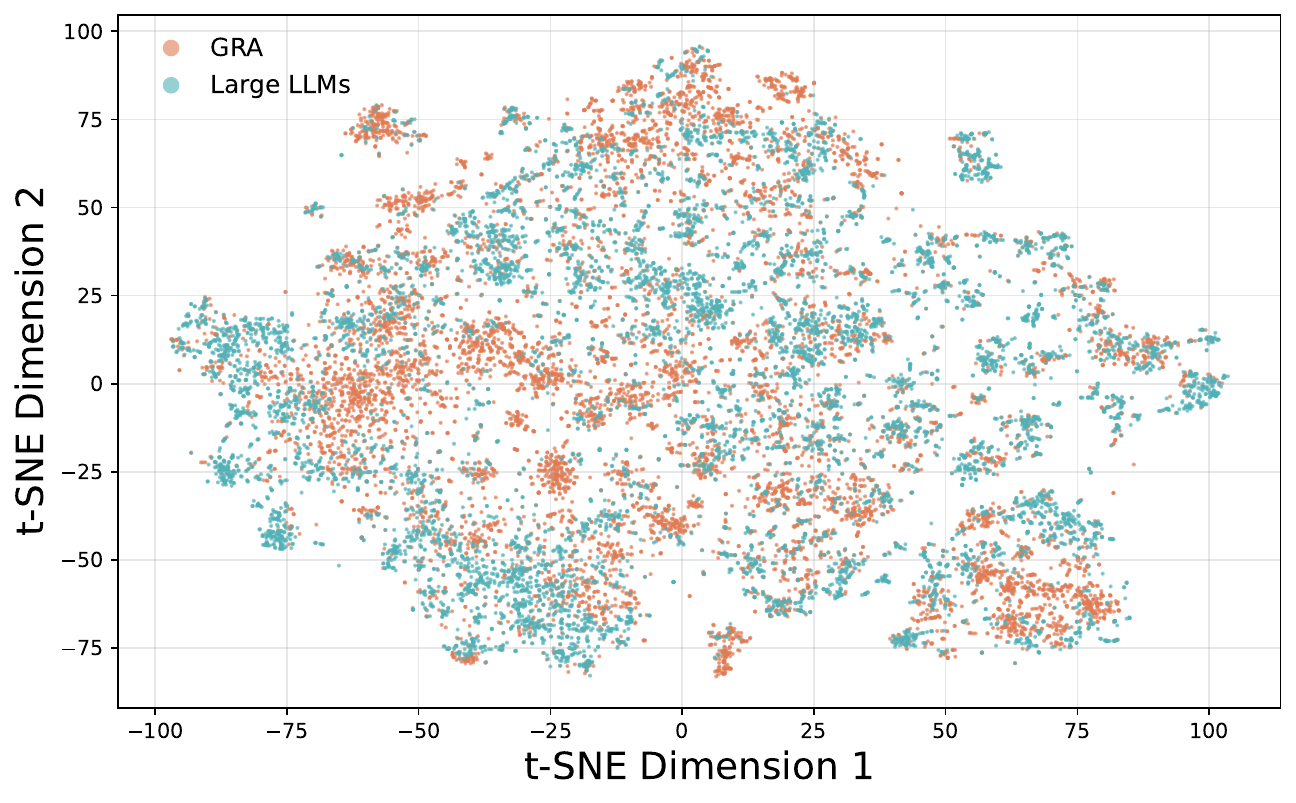}
        \vspace{-0.4cm}
        \caption{Data coverage comparison between large LLMs distilled data and GRA synthetic data.}
        \vspace{-1.0cm}
        \label{fig:tsne_bigllm_GRA}
    \end{minipage}
\end{figure}

\vspace{-0.2cm}
\subsection{Data Diversity}
\vspace{-0.1cm}
To assess the diversity of synthesized data, we compare three sources: GRA-generated data, data distilled from a single large LLM, and the vanilla seed dataset. We visualize the instruction embeddings with t-SNE~\cite{van2008tsne}.
\paragraph{Diversity Analysis.} As shown in Figure~\ref{fig:tsne_seed_GRA}. GRA-generated data exhibits a broader distribution than the vanilla seed dataset, particularly covering underrepresented regions in the seed data’s embedding space. This suggests that GRA complementing the seed dataset by expanding its coverage. Besides, Figure~\ref{fig:tsne_bigllm_GRA} reveals that GRA surpasses the single large LLM in distribution diversity, implying that collaborative small LLMs can mitigate the inherent limits and biases of single model. Notably, while the seed dataset demonstrates comparable instruction diversity, its inferior performance in main experiments-Table \ref{tab:main_result} highlights critical deficiencies in response quality rather than instruction coverage.

\vspace{-0.1cm}
\paragraph{Seed Data Enhancement via GRA.} We further applied GRA’s generator to rewrite responses in the vanilla seed dataset, followed by curation using GRA’s reviewer and adjudicator. As shown in Table \ref{tab:Refine}, the refined dataset achieves performance parity with GRA-generated data. Moreover, combining GRA-synthesized data with enhanced seed data yields synergistic effects, outperforming either dataset alone in certain settings. This underscores GRA’s dual utility: (a) generating novel high-quality data and (b) upgrading existing data through systematic rewriting and quality control.

\begin{table}[htbp]
\renewcommand{\arraystretch}{1.15}
\resizebox{\linewidth}{!}{
\begin{tabular}{cccc}
\hline
Data~/~Seed Dataset                     & Alpaca & WizardLM & Condor \\ \hline
\multicolumn{4}{c}{Llama-3.1-8B-Base}                                \\ \hline
Vanilla Seed Data                       & 45.88  & 46.59    & 43.04  \\
GRA-Refined Seed Data                   & 50.39  & 52.81    & 47.68  \\
GRA Synthetic Data                      & 52.14  & 50.74    & 51.16  \\
GRA Integrated Data                     & 51.93  & 52.35    & 52.22  \\ \hline
\multicolumn{4}{c}{Qwen-2.5-7B-Base}                                 \\ \hline
Vanilla Seed Data                       & 49.32  & 45.71    & 50.61  \\
GRA-Refined Seed Data                   & 59.97  & 58.34    & 45.5   \\
GRA Synthetic Data                      & 59.52  & 62.17    & 61.13  \\
GRA Integrated Data & 61.19  & 59.00       & 62.1   \\ \hline
\end{tabular}}
\vspace{-0.2cm}
\caption{Comparison across Vanilla Seed Data, GRA-Refined Seed Data, GRA Synthetic Data and GRA Integrated Data.}
\label{tab:Refine}
\end{table}
\vspace{-0.7cm}

\subsection{Data Quality}
\label{sec:analysis_quality}
\begin{figure}[!t]
    \centering
    \includegraphics[width=0.9\linewidth]{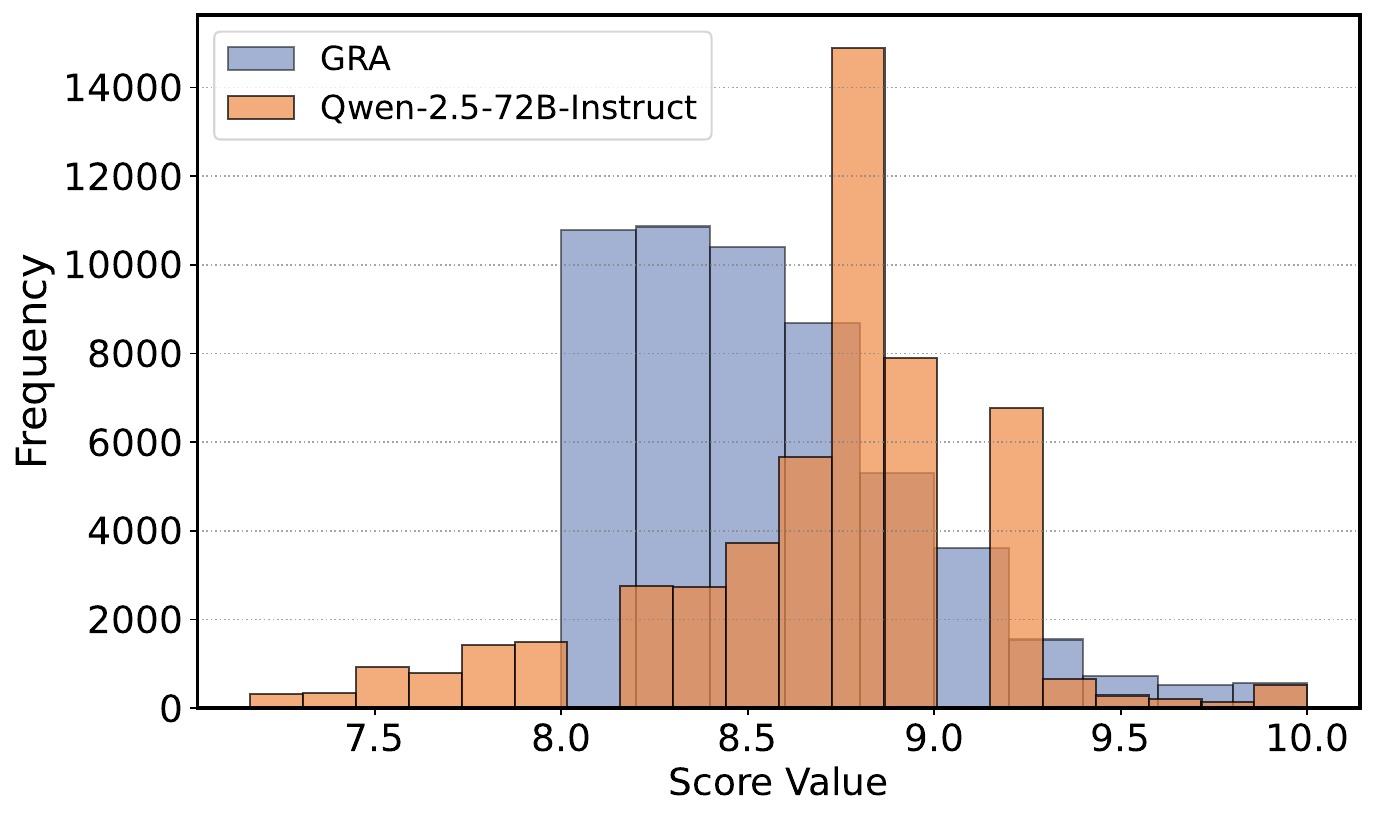}
    \vspace{-0.5cm}
    \caption{Data quality score  comparison between GRA and single Large LLMs—Qwen-2.5-72B-Instruct.}
    \vspace{-0.4cm}
    \label{fig:score_compare}
\end{figure}

To evaluate the data quality, we utilize Qwen-2.5-72B-Instruct to assess the synthetic data and compare its score with our Reviewers' score. 
From Figure~\ref{fig:score_compare}, we find that 
the GRA method exhibits finer granularity and a more balanced scoring distribution compared to the coarser-grained scores from the large LLM, indicating its ability to capture nuanced quality variations. Second, over $87.3$\% of GRA-curated data (scores $> 8$) align with high scores from the large LLM, while $12.7$\% receive lower scores ($\leq8$) from the latter. Notably, the large LLM assigns even higher scores to a larger subset of data, further validating the reliability of GRA in generating high-quality data. These results suggest that GRA enhances scoring resolution without compromising assessment consistency with large LLMs.

\vspace{-0.1cm}
\subsection{Data Difficulty} 
\begin{figure}[!t]
    \centering
    \includegraphics[width=0.9\linewidth]{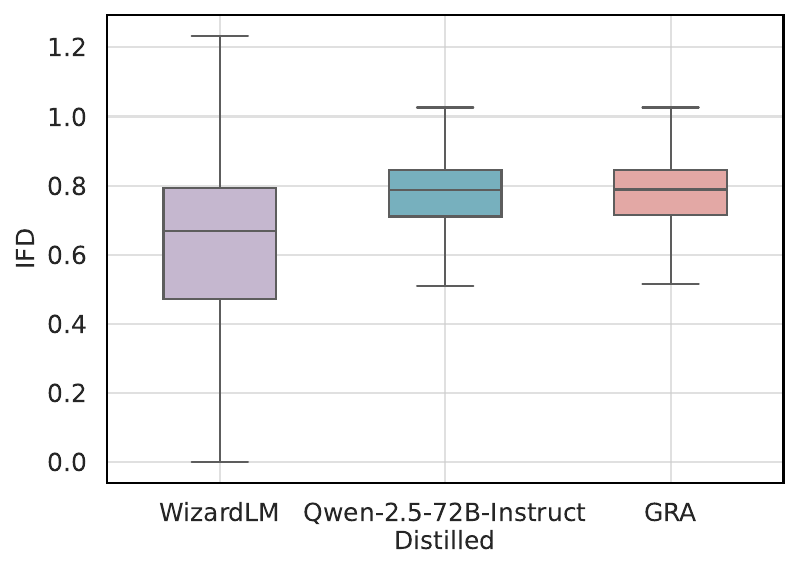}
    \vspace{-0.4cm}
    \caption{The IFD score comparison between the data generated by GRA, the vanilla seed dataset WizardLM, and single large LLM (Qwen-2.5-72B-Instruct) distilled data.}
    \vspace{-0.7cm}
    \label{fig:analysis_ifd}
\end{figure}

\vspace{-0.1cm}
To evaluate the complexity and knowledge density of synthesized data, we compute the Instruction-Following Difficulty (IFD) score~\cite{ifd} for the three datasets.
From the results depicted in Figure~\ref{fig:analysis_ifd}, two key observations emerge: 
First, GRA-generated data achieves a significantly higher IFD than the seed dataset ($+14.58$\%), indicating its enhanced difficulty and richer knowledge coverage. This aligns with empirical training results, where models trained on GRA data exhibit superior performance. 
Second, despite using smaller LLMs, GRA attains comparable difficulty to the Qwen-72B-distilled data ($75.82$\% v.s. $75.49$\%). This suggests that collaborative small LLMs in GRA mitigate individual capability limitations, effectively emulating large LLMs in generating challenging examples. 
These results validate GRA’s ability to synthesize high-quality, difficulty-controllable data without relying on large LLMs, addressing both scalability and performance trade-offs in data synthesis pipelines.

\vspace{-0.2cm}
\section{Conclusion}
\vspace{-0.3cm}
In this paper, we introduce GRA (Generator-Reviewer-Adjudicator), a novel framework that leverages the collaborative capabilities of multiple small LLMs to achieve data synthesis quality comparable to that of single monolithic LLMs. Our approach addresses the significant computational and environmental concerns associated with large LLMs while mitigating the biases inherent in their centralized architectures.
The experimental results across diverse benchmarks validate the efficacy of GRA. GRA underscores the potential of role specialization in enhancing data synthesis quality, challenging the prevailing notion that larger models are inherently superior for such tasks, paving the way for more sustainable and equitable data synthesis.

\clearpage
\section*{Limitations}
Despite the promising performance of the GRA framework, several aspects warrant further exploration. Current role allocation mechanism—based on randomized assignment—may yield suboptimal collaboration patterns. This stochastic approach lacks explicit optimization for task-specific synergies, suggesting the need for intelligent role configuration strategies (e.g., performance-aware dynamic allocation or metric-driven reinforcement learning) to systematically maximize agent complementarity. Furthermore, while GRA is primarily designed for textual reasoning tasks, extending its paradigm to multimodal scenarios presents a compelling direction.

\bibliography{custom}

\begin{thebibliography}{52}
\providecommand{\natexlab}[1]{#1}

\bibitem[{Ahn et~al.(2024)Ahn, Verma, Lou, Liu, Zhang, and Yin}]{llm_math_survey}
Janice Ahn, Rishu Verma, Renze Lou, Di~Liu, Rui Zhang, and Wenpeng Yin. 2024.
\newblock Large language models for mathematical reasoning: Progresses and challenges.
\newblock In \emph{{EACL} (Student Research Workshop)}, pages 225--237. Association for Computational Linguistics.

\bibitem[{Austin et~al.(2021)Austin, Odena, Nye, Bosma, Michalewski, Dohan, Jiang, Cai, Terry, Le et~al.}]{austin2021program}
Jacob Austin, Augustus Odena, Maxwell Nye, Maarten Bosma, Henryk Michalewski, David Dohan, Ellen Jiang, Carrie Cai, Michael Terry, Quoc Le, et~al. 2021.
\newblock Program synthesis with large language models.
\newblock \emph{arXiv preprint arXiv:2108.07732}.

\bibitem[{Cai et~al.(2024)Cai, Cao, Chen, Chen, Chen, Chen, Chen, Chen, Chen, Chu, Dong, Duan, Fan, Fei, Gao, Ge, Gu, Gu, Gui, Guo, Guo, He, Hu, Huang, Jiang, Jiao, Jin, Lei, Li, Li, Li, Li, Li, Li, Liu, Liu, Hong, Liu, Liu, Liu, Lv, Lv, Lv, Ma, Ma, Ma, Ning, Ouyang, Qiu, Qu, Shang, Shao, Song, Song, Sui, Sun, Sun, Tang, Wang, Wang, Wang, Wang, Wang, Wang, Wang, Wei, Weng, Wu, Xiong, Xu, Xu, Yan, Yan, Yang, Ye, Ying, Yu, Yu, Zang, Zhang, Zhang, Zhang, Zhang, Zhang, Zhang, Zhang, Zhang, Zhang, Zhang, Zhang, Zhao, Zhao, Zhao, Zhou, Zhou, Zhuo, Zou, Qiu, Qiao, and Lin}]{cai2024internlm2}
Zheng Cai, Maosong Cao, Haojiong Chen, Kai Chen, Keyu Chen, Xin Chen, Xun Chen, Zehui Chen, Zhi Chen, Pei Chu, Xiaoyi Dong, Haodong Duan, Qi~Fan, Zhaoye Fei, Yang Gao, Jiaye Ge, Chenya Gu, Yuzhe Gu, Tao Gui, Aijia Guo, Qipeng Guo, Conghui He, Yingfan Hu, Ting Huang, Tao Jiang, Penglong Jiao, Zhenjiang Jin, Zhikai Lei, Jiaxing Li, Jingwen Li, Linyang Li, Shuaibin Li, Wei Li, Yining Li, Hongwei Liu, Jiangning Liu, Jiawei Hong, Kaiwen Liu, Kuikun Liu, Xiaoran Liu, Chengqi Lv, Haijun Lv, Kai Lv, Li~Ma, Runyuan Ma, Zerun Ma, Wenchang Ning, Linke Ouyang, Jiantao Qiu, Yuan Qu, Fukai Shang, Yunfan Shao, Demin Song, Zifan Song, Zhihao Sui, Peng Sun, Yu~Sun, Huanze Tang, Bin Wang, Guoteng Wang, Jiaqi Wang, Jiayu Wang, Rui Wang, Yudong Wang, Ziyi Wang, Xingjian Wei, Qizhen Weng, Fan Wu, Yingtong Xiong, Chao Xu, Ruiliang Xu, Hang Yan, Yirong Yan, Xiaogui Yang, Haochen Ye, Huaiyuan Ying, Jia Yu, Jing Yu, Yuhang Zang, Chuyu Zhang, Li~Zhang, Pan Zhang, Peng Zhang, Ruijie Zhang, Shuo Zhang, Songyang Zhang, Wenjian Zhang,
  Wenwei Zhang, Xingcheng Zhang, Xinyue Zhang, Hui Zhao, Qian Zhao, Xiaomeng Zhao, Fengzhe Zhou, Zaida Zhou, Jingming Zhuo, Yicheng Zou, Xipeng Qiu, Yu~Qiao, and Dahua Lin. 2024.
\newblock \href {https://arxiv.org/abs/2403.17297} {Internlm2 technical report}.
\newblock \emph{Preprint}, arXiv:2403.17297.

\bibitem[{Cao et~al.(2025)Cao, Zhang, Li, Zhang, Liu, Duan, Zhang, and Chen}]{cao2025condor}
Maosong Cao, Taolin Zhang, Mo~Li, Chuyu Zhang, Yunxin Liu, Haodong Duan, Songyang Zhang, and Kai Chen. 2025.
\newblock Condor: Enhance llm alignment with knowledge-driven data synthesis and refinement.
\newblock \emph{arXiv preprint arXiv:2501.12273}.

\bibitem[{Chen and Varoquaux(2024)}]{chen2024role}
Lihu Chen and Ga{\"e}l Varoquaux. 2024.
\newblock What is the role of small models in the llm era: A survey.
\newblock \emph{arXiv preprint arXiv:2409.06857}.

\bibitem[{Chen et~al.(2021)Chen, Tworek, Jun, Yuan, Pinto, Kaplan, Edwards, Burda, Joseph, Brockman et~al.}]{chen2021evaluating}
Mark Chen, Jerry Tworek, Heewoo Jun, Qiming Yuan, Henrique Ponde De~Oliveira Pinto, Jared Kaplan, Harri Edwards, Yuri Burda, Nicholas Joseph, Greg Brockman, et~al. 2021.
\newblock Evaluating large language models trained on code.
\newblock \emph{arXiv preprint arXiv:2107.03374}.

\bibitem[{Clark et~al.(2018)Clark, Cowhey, Etzioni, Khot, Sabharwal, Schoenick, and Tafjord}]{clark2018think}
Peter Clark, Isaac Cowhey, Oren Etzioni, Tushar Khot, Ashish Sabharwal, Carissa Schoenick, and Oyvind Tafjord. 2018.
\newblock Think you have solved question answering? try arc, the ai2 reasoning challenge.
\newblock \emph{arXiv preprint arXiv:1803.05457}.

\bibitem[{Cobbe et~al.(2021)Cobbe, Kosaraju, Bavarian, Chen, Jun, Kaiser, Plappert, Tworek, Hilton, Nakano et~al.}]{cobbe2021gsm8k}
Karl Cobbe, Vineet Kosaraju, Mohammad Bavarian, Mark Chen, Heewoo Jun, Lukasz Kaiser, Matthias Plappert, Jerry Tworek, Jacob Hilton, Reiichiro Nakano, et~al. 2021.
\newblock Training verifiers to solve math word problems.
\newblock \emph{arXiv preprint arXiv:2110.14168}.

\bibitem[{Contributors(2023{\natexlab{a}})}]{2023lmdeploy}
LMDeploy Contributors. 2023{\natexlab{a}}.
\newblock Lmdeploy: A toolkit for compressing, deploying, and serving llm.
\newblock \url{https://github.com/InternLM/lmdeploy}.

\bibitem[{Contributors(2023{\natexlab{b}})}]{2023opencompass}
OpenCompass Contributors. 2023{\natexlab{b}}.
\newblock Opencompass: A universal evaluation platform for foundation models.
\newblock \url{https://github.com/open-compass/opencompass}.

\bibitem[{Dubey et~al.(2024{\natexlab{a}})Dubey, Jauhri, Pandey, Kadian, Al-Dahle, Letman, Mathur, Schelten, Yang, Fan et~al.}]{llama3}
Abhimanyu Dubey, Abhinav Jauhri, Abhinav Pandey, Abhishek Kadian, Ahmad Al-Dahle, Aiesha Letman, Akhil Mathur, Alan Schelten, Amy Yang, Angela Fan, et~al. 2024{\natexlab{a}}.
\newblock The llama 3 herd of models.
\newblock \emph{arXiv preprint arXiv:2407.21783}.

\bibitem[{Dubey et~al.(2024{\natexlab{b}})Dubey, Jauhri, Pandey, Kadian, Al-Dahle, Letman, Mathur, Schelten, Yang, Fan et~al.}]{dubey2024llama}
Abhimanyu Dubey, Abhinav Jauhri, Abhinav Pandey, Abhishek Kadian, Ahmad Al-Dahle, Aiesha Letman, Akhil Mathur, Alan Schelten, Amy Yang, Angela Fan, et~al. 2024{\natexlab{b}}.
\newblock The llama 3 herd of models.
\newblock \emph{arXiv preprint arXiv:2407.21783}.

\bibitem[{Gu et~al.(2024)Gu, Jiang, Shi, Tan, Zhai, Xu, Li, Shen, Ma, Liu et~al.}]{gu2024survey}
Jiawei Gu, Xuhui Jiang, Zhichao Shi, Hexiang Tan, Xuehao Zhai, Chengjin Xu, Wei Li, Yinghan Shen, Shengjie Ma, Honghao Liu, et~al. 2024.
\newblock A survey on llm-as-a-judge.
\newblock \emph{arXiv preprint arXiv:2411.15594}.

\bibitem[{Guo et~al.(2025)Guo, Yang, Zhang, Song, Zhang, Xu, Zhu, Ma, Wang, Bi et~al.}]{deepseek_r1}
Daya Guo, Dejian Yang, Haowei Zhang, Junxiao Song, Ruoyu Zhang, Runxin Xu, Qihao Zhu, Shirong Ma, Peiyi Wang, Xiao Bi, et~al. 2025.
\newblock Deepseek-r1: Incentivizing reasoning capability in llms via reinforcement learning.
\newblock \emph{arXiv preprint arXiv:2501.12948}.

\bibitem[{Hagos et~al.(2024)Hagos, Battle, and Rawat}]{hagos2024recent}
Desta~Haileselassie Hagos, Rick Battle, and Danda~B Rawat. 2024.
\newblock Recent advances in generative ai and large language models: Current status, challenges, and perspectives.
\newblock \emph{IEEE Transactions on Artificial Intelligence}.

\bibitem[{Hendrycks et~al.(2020)Hendrycks, Burns, Basart, Zou, Mazeika, Song, and Steinhardt}]{hendrycks2020measuring}
Dan Hendrycks, Collin Burns, Steven Basart, Andy Zou, Mantas Mazeika, Dawn Song, and Jacob Steinhardt. 2020.
\newblock Measuring massive multitask language understanding.
\newblock \emph{arXiv preprint arXiv:2009.03300}.

\bibitem[{Hendrycks et~al.(2021)Hendrycks, Burns, Kadavath, Arora, Basart, Tang, Song, and Steinhardt}]{hendrycks2021math}
Dan Hendrycks, Collin Burns, Saurav Kadavath, Akul Arora, Steven Basart, Eric Tang, Dawn Song, and Jacob Steinhardt. 2021.
\newblock \href {https://datasets-benchmarks-proceedings.neurips.cc/paper_files/paper/2021/file/be83ab3ecd0db773eb2dc1b0a17836a1-Paper-round2.pdf} {Measuring mathematical problem solving with the math dataset}.
\newblock In \emph{Proceedings of the Neural Information Processing Systems Track on Datasets and Benchmarks}, volume~1.

\bibitem[{Hinton(2015)}]{hinton2015distilling}
Geoffrey Hinton. 2015.
\newblock Distilling the knowledge in a neural network.
\newblock \emph{arXiv preprint arXiv:1503.02531}.

\bibitem[{Jiang et~al.(2023)Jiang, Sablayrolles, Mensch, Bamford, Chaplot, Casas, Bressand, Lengyel, Lample, Saulnier et~al.}]{jiang2023mistral}
Albert~Q Jiang, Alexandre Sablayrolles, Arthur Mensch, Chris Bamford, Devendra~Singh Chaplot, Diego de~las Casas, Florian Bressand, Gianna Lengyel, Guillaume Lample, Lucile Saulnier, et~al. 2023.
\newblock Mistral 7b.
\newblock \emph{arXiv preprint arXiv:2310.06825}.

\bibitem[{Jiang et~al.(2024)Jiang, Wang, Shen, Kim, and Kim}]{llm_codegen_survey}
Juyong Jiang, Fan Wang, Jiasi Shen, Sungju Kim, and Sunghun Kim. 2024.
\newblock A survey on large language models for code generation.
\newblock \emph{arXiv preprint arXiv:2406.00515}.

\bibitem[{Kwon et~al.(2023)Kwon, Li, Zhuang, Sheng, Zheng, Yu, Gonzalez, Zhang, and Stoica}]{kwon2023efficient}
Woosuk Kwon, Zhuohan Li, Siyuan Zhuang, Ying Sheng, Lianmin Zheng, Cody~Hao Yu, Joseph Gonzalez, Hao Zhang, and Ion Stoica. 2023.
\newblock Efficient memory management for large language model serving with pagedattention.
\newblock In \emph{Proceedings of the 29th Symposium on Operating Systems Principles}, pages 611--626.

\bibitem[{Lambert et~al.(2024)Lambert, Morrison, Pyatkin, Huang, Ivison, Brahman, Miranda, Liu, Dziri, Lyu et~al.}]{lambert2024t}
Nathan Lambert, Jacob Morrison, Valentina Pyatkin, Shengyi Huang, Hamish Ivison, Faeze Brahman, Lester James~V Miranda, Alisa Liu, Nouha Dziri, Shane Lyu, et~al. 2024.
\newblock T$\backslash$" ulu 3: Pushing frontiers in open language model post-training.
\newblock \emph{arXiv preprint arXiv:2411.15124}.

\bibitem[{Li et~al.(2024{\natexlab{a}})Li, Yuan, Yuan, Dong, Lu, Wu, Tan, Wang, and Zhou}]{muggle_math}
Chengpeng Li, Zheng Yuan, Hongyi Yuan, Guanting Dong, Keming Lu, Jiancan Wu, Chuanqi Tan, Xiang Wang, and Chang Zhou. 2024{\natexlab{a}}.
\newblock Mugglemath: Assessing the impact of query and response augmentation on math reasoning.
\newblock In \emph{{ACL} {(1)}}, pages 10230--10258. Association for Computational Linguistics.

\bibitem[{Li et~al.(2016)Li, Galley, Brockett, Gao, and Dolan}]{ngram}
Jiwei Li, Michel Galley, Chris Brockett, Jianfeng Gao, and Bill Dolan. 2016.
\newblock A diversity-promoting objective function for neural conversation models.
\newblock In \emph{{HLT-NAACL}}, pages 110--119. The Association for Computational Linguistics.

\bibitem[{Li et~al.(2024{\natexlab{b}})Li, Zhang, Li, Chen, Chen, Cheng, Wang, Zhou, and Xiao}]{ifd}
Ming Li, Yong Zhang, Zhitao Li, Jiuhai Chen, Lichang Chen, Ning Cheng, Jianzong Wang, Tianyi Zhou, and Jing Xiao. 2024{\natexlab{b}}.
\newblock From quantity to quality: Boosting {LLM} performance with self-guided data selection for instruction tuning.
\newblock In \emph{{NAACL-HLT}}, pages 7602--7635. Association for Computational Linguistics.

\bibitem[{Liu et~al.(2024)Liu, Feng, Xue, Wang, Wu, Lu, Zhao, Deng, Zhang, Ruan et~al.}]{deepseek_v3}
Aixin Liu, Bei Feng, Bing Xue, Bingxuan Wang, Bochao Wu, Chengda Lu, Chenggang Zhao, Chengqi Deng, Chenyu Zhang, Chong Ruan, et~al. 2024.
\newblock Deepseek-v3 technical report.
\newblock \emph{arXiv preprint arXiv:2412.19437}.

\bibitem[{Long et~al.(2024)Long, Wang, Xiao, Zhao, Ding, Chen, and Wang}]{long2024llms}
Lin Long, Rui Wang, Ruixuan Xiao, Junbo Zhao, Xiao Ding, Gang Chen, and Haobo Wang. 2024.
\newblock On llms-driven synthetic data generation, curation, and evaluation: {A} survey.
\newblock In \emph{{ACL} (Findings)}, pages 11065--11082. Association for Computational Linguistics.

\bibitem[{Majumdar et~al.(2024)Majumdar, Noroozi, Narenthiran, Ficek, Balam, and Ginsburg}]{geneticinstruct}
Somshubra Majumdar, Vahid Noroozi, Sean Narenthiran, Aleksander Ficek, Jagadeesh Balam, and Boris Ginsburg. 2024.
\newblock Genetic instruct: Scaling up synthetic generation of coding instructions for large language models.
\newblock \emph{arXiv preprint arXiv:2407.21077}.

\bibitem[{Mitra et~al.(2024)Mitra, Del~Corro, Zheng, Mahajan, Rouhana, Codas, Lu, Chen, Vrousgos, Rosset et~al.}]{agentinstruct}
Arindam Mitra, Luciano Del~Corro, Guoqing Zheng, Shweti Mahajan, Dany Rouhana, Andres Codas, Yadong Lu, Wei-ge Chen, Olga Vrousgos, Corby Rosset, et~al. 2024.
\newblock Agentinstruct: Toward generative teaching with agentic flows.
\newblock \emph{arXiv preprint arXiv:2407.03502}.

\bibitem[{Naveed et~al.(2023)Naveed, Khan, Qiu, Saqib, Anwar, Usman, Akhtar, Barnes, and Mian}]{naveed2023comprehensive}
Humza Naveed, Asad~Ullah Khan, Shi Qiu, Muhammad Saqib, Saeed Anwar, Muhammad Usman, Naveed Akhtar, Nick Barnes, and Ajmal Mian. 2023.
\newblock A comprehensive overview of large language models.
\newblock \emph{arXiv preprint arXiv:2307.06435}.

\bibitem[{OpenAI(2024)}]{openai_o1}
OpenAI. 2024.
\newblock \href {https://openai.com/index/learning-to-reason-with-llms} {Learning to reason with llms}.

\bibitem[{OpenAI et~al.(2023)OpenAI, Adler, Agarwal, Ahmad, Akkaya, Aleman, Almeida, Altenschmidt, Altman, Anadkat et~al.}]{openai_gpt4}
Josh OpenAI, Achiam, Steven Adler, Sandhini Agarwal, Lama Ahmad, Ilge Akkaya, Florencia~Leoni Aleman, Diogo Almeida, Janko Altenschmidt, Sam Altman, Shyamal Anadkat, et~al. 2023.
\newblock Gpt-4 technical report.
\newblock \emph{arXiv preprint arXiv:2303.08774}.

\bibitem[{Reimers and Gurevych(2019)}]{reimers-2019-sentence-bert}
Nils Reimers and Iryna Gurevych. 2019.
\newblock \href {https://arxiv.org/abs/1908.10084} {Sentence-bert: Sentence embeddings using siamese bert-networks}.
\newblock In \emph{Proceedings of the 2019 Conference on Empirical Methods in Natural Language Processing}. Association for Computational Linguistics.

\bibitem[{Rein et~al.(2023)Rein, Hou, Stickland, Petty, Pang, Dirani, Michael, and Bowman}]{rein2023gpqa}
David Rein, Betty~Li Hou, Asa~Cooper Stickland, Jackson Petty, Richard~Yuanzhe Pang, Julien Dirani, Julian Michael, and Samuel~R Bowman. 2023.
\newblock Gpqa: A graduate-level google-proof q\&a benchmark.
\newblock \emph{arXiv preprint arXiv:2311.12022}.

\bibitem[{Salkar et~al.(2022)Salkar, Trikalinos, Wallace, and Nenkova}]{self_repetition}
Nikita Salkar, Thomas~A. Trikalinos, Byron~C. Wallace, and Ani Nenkova. 2022.
\newblock Self-repetition in abstractive neural summarizers.
\newblock In \emph{{AACL/IJCNLP} {(2)}}, pages 341--350. Association for Computational Linguistics.

\bibitem[{Shu et~al.(2019)Shu, Nakayama, and Cho}]{shu2019generating}
Raphael Shu, Hideki Nakayama, and Kyunghyun Cho. 2019.
\newblock Generating diverse translations with sentence codes.
\newblock In \emph{{ACL} {(1)}}, pages 1823--1827. Association for Computational Linguistics.

\bibitem[{Suzgun et~al.(2022)Suzgun, Scales, Sch{\"a}rli, Gehrmann, Tay, Chung, Chowdhery, Le, Chi, Zhou et~al.}]{suzgun2022challenging}
Mirac Suzgun, Nathan Scales, Nathanael Sch{\"a}rli, Sebastian Gehrmann, Yi~Tay, Hyung~Won Chung, Aakanksha Chowdhery, Quoc~V Le, Ed~H Chi, Denny Zhou, et~al. 2022.
\newblock Challenging big-bench tasks and whether chain-of-thought can solve them.
\newblock \emph{arXiv preprint arXiv:2210.09261}.

\bibitem[{Taori et~al.(2023)Taori, Gulrajani, Zhang, Dubois, Li, Guestrin, Liang, and Hashimoto}]{alpaca}
Rohan Taori, Ishaan Gulrajani, Tianyi Zhang, Yann Dubois, Xuechen Li, Carlos Guestrin, Percy Liang, and Tatsunori~B. Hashimoto. 2023.
\newblock Stanford alpaca: An instruction-following llama model.
\newblock \url{https://github.com/tatsu-lab/stanford_alpaca}.

\bibitem[{Van~der Maaten and Hinton(2008)}]{van2008tsne}
Laurens Van~der Maaten and Geoffrey Hinton. 2008.
\newblock Visualizing data using t-sne.
\newblock \emph{Journal of machine learning research}, 9(11).

\bibitem[{Wang et~al.(2024)Wang, Zhu, Ren, Liu, Li, Zhang, Zhang, Wu, Zhan, Liu et~al.}]{wang2024survey}
Ke~Wang, Jiahui Zhu, Minjie Ren, Zeming Liu, Shiwei Li, Zongye Zhang, Chenkai Zhang, Xiaoyu Wu, Qiqi Zhan, Qingjie Liu, et~al. 2024.
\newblock A survey on data synthesis and augmentation for large language models.
\newblock \emph{arXiv preprint arXiv:2410.12896}.

\bibitem[{Xu et~al.(2024{\natexlab{a}})Xu, Sun, Zheng, Geng, Zhao, Feng, Tao, Lin, and Jiang}]{xu2023wizardlm}
Can Xu, Qingfeng Sun, Kai Zheng, Xiubo Geng, Pu~Zhao, Jiazhan Feng, Chongyang Tao, Qingwei Lin, and Daxin Jiang. 2024{\natexlab{a}}.
\newblock Wizardlm: Empowering large pre-trained language models to follow complex instructions.
\newblock In \emph{{ICLR}}. OpenReview.net.

\bibitem[{Xu et~al.(2024{\natexlab{b}})Xu, Li, Tao, Shen, Cheng, Li, Xu, Tao, and Zhou}]{xu2024survey}
Xiaohan Xu, Ming Li, Chongyang Tao, Tao Shen, Reynold Cheng, Jinyang Li, Can Xu, Dacheng Tao, and Tianyi Zhou. 2024{\natexlab{b}}.
\newblock A survey on knowledge distillation of large language models.
\newblock \emph{arXiv preprint arXiv:2402.13116}.

\bibitem[{Xu et~al.(2025)Xu, Jiang, Niu, Deng, Poovendran, Choi, and Lin}]{magpie}
Zhangchen Xu, Fengqing Jiang, Luyao Niu, Yuntian Deng, Radha Poovendran, Yejin Choi, and Bill~Yuchen Lin. 2025.
\newblock \href {https://openreview.net/forum?id=Pnk7vMbznK} {Magpie: Alignment data synthesis from scratch by prompting aligned {LLM}s with nothing}.
\newblock In \emph{The Thirteenth International Conference on Learning Representations}.

\bibitem[{Yang et~al.(2024{\natexlab{a}})Yang, Yang, Zhang, Hui, Zheng, Yu, Li, Liu, Huang, Wei et~al.}]{qwen25}
An~Yang, Baosong Yang, Beichen Zhang, Binyuan Hui, Bo~Zheng, Bowen Yu, Chengyuan Li, Dayiheng Liu, Fei Huang, Haoran Wei, et~al. 2024{\natexlab{a}}.
\newblock Qwen2. 5 technical report.
\newblock \emph{arXiv preprint arXiv:2412.15115}.

\bibitem[{Yang et~al.(2024{\natexlab{b}})Yang, Yang, Zhang, Hui, Zheng, Yu, Li, Liu, Huang, Wei et~al.}]{yang2024qwen2}
An~Yang, Baosong Yang, Beichen Zhang, Binyuan Hui, Bo~Zheng, Bowen Yu, Chengyuan Li, Dayiheng Liu, Fei Huang, Haoran Wei, et~al. 2024{\natexlab{b}}.
\newblock Qwen2. 5 technical report.
\newblock \emph{arXiv preprint arXiv:2412.15115}.

\bibitem[{Yao et~al.(2024)Yao, Duan, Xu, Cai, Sun, and Zhang}]{yao2024survey}
Yifan Yao, Jinhao Duan, Kaidi Xu, Yuanfang Cai, Zhibo Sun, and Yue Zhang. 2024.
\newblock A survey on large language model (llm) security and privacy: The good, the bad, and the ugly.
\newblock \emph{High-Confidence Computing}, page 100211.

\bibitem[{Zellers et~al.(2019)Zellers, Holtzman, Bisk, Farhadi, and Choi}]{zellers2019hellaswag}
Rowan Zellers, Ari Holtzman, Yonatan Bisk, Ali Farhadi, and Yejin Choi. 2019.
\newblock Hellaswag: Can a machine really finish your sentence?
\newblock \emph{arXiv preprint arXiv:1905.07830}.

\bibitem[{Zhao et~al.(2023)Zhao, Zhou, Li, Tang, Wang, Hou, Min, Zhang, Zhang, Dong et~al.}]{zhao2023survey}
Wayne~Xin Zhao, Kun Zhou, Junyi Li, Tianyi Tang, Xiaolei Wang, Yupeng Hou, Yingqian Min, Beichen Zhang, Junjie Zhang, Zican Dong, et~al. 2023.
\newblock A survey of large language models.
\newblock \emph{arXiv preprint arXiv:2303.18223}.

\bibitem[{Zheng et~al.(2024{\natexlab{a}})Zheng, Zhang, Zhang, Ye, Luo, Feng, and Ma}]{zheng2024llamafactory}
Yaowei Zheng, Richong Zhang, Junhao Zhang, Yanhan Ye, Zheyan Luo, Zhangchi Feng, and Yongqiang Ma. 2024{\natexlab{a}}.
\newblock Llamafactory: Unified efficient fine-tuning of 100+ language models.
\newblock \emph{arXiv preprint arXiv:2403.13372}.

\bibitem[{Zheng et~al.(2024{\natexlab{b}})Zheng, Chen, Qian, Shi, Shu, and Chen}]{zheng2024review}
Yue Zheng, Yuhao Chen, Bin Qian, Xiufang Shi, Yuanchao Shu, and Jiming Chen. 2024{\natexlab{b}}.
\newblock A review on edge large language models: Design, execution, and applications.
\newblock \emph{arXiv preprint arXiv:2410.11845}.

\bibitem[{Zhou et~al.(2023)Zhou, Lu, Mishra, Brahma, Basu, Luan, Zhou, and Hou}]{zhou2023instruction}
Jeffrey Zhou, Tianjian Lu, Swaroop Mishra, Siddhartha Brahma, Sujoy Basu, Yi~Luan, Denny Zhou, and Le~Hou. 2023.
\newblock Instruction-following evaluation for large language models.
\newblock \emph{arXiv preprint arXiv:2311.07911}.

\bibitem[{Zhu et~al.(2018)Zhu, Lu, Zheng, Guo, Zhang, Wang, and Yu}]{texygen}
Yaoming Zhu, Sidi Lu, Lei Zheng, Jiaxian Guo, Weinan Zhang, Jun Wang, and Yong Yu. 2018.
\newblock Texygen: {A} benchmarking platform for text generation models.
\newblock In \emph{{SIGIR}}, pages 1097--1100. {ACM}.

\end{thebibliography}

\appendix

\section{Generation}
\label{sec:app_gen}
When generating data, we deployed models using VLLM and LMDeploy~\cite{2023lmdeploy} on 8 NVIDIA A100 GPUs. For setting the model generation parameters, we used the default settings of the deployment tools, except for temperature=0.2, top\_p=0.9,  max\_token=4096. The number of examples used in few-shot ranges from $2$ to $4$. We set the minimum score threshold $\tau = 8$ and the maximum standard deviation $\delta = 1.5$ at review stage. For deduplication, the maximum cosine similarity is set to $0.9$. 
Here we provide detailed criteria for data domain classification:
\begin{itemize}
    \item \textbf{Coding}: Involves understanding, generating, debugging, and fixing code.
    \item \textbf{Math}: Entails performing mathematical operations, applying formulas, and solving problems.
    \item \textbf{QA}: Focuses on providing specialized and accurate answers to questions within a specific domain or field.
    \item \textbf{Language}: Involves accurately understanding and analyzing text, performing tasks such as translation, summarization, and classification.
    \item \textbf{Role Play}: Engages in simulated conversations or behaviors by assuming different roles to explore various perspectives or scenarios.
    \item \textbf{Reasoning}: Involves complex causal reasoning and logical judgment, solving multi-level problems through inference and analysis.
    \item \textbf{Creation}: Emphasizes originality and creativity, creating content according to requirements and adapting to different styles and themes.
\end{itemize}
Before initiating the GRA data synthesis, we employed all available LLMs to extract keywords and summaries from the seed dataset using a randomized rotation approach.

\section{Training}
\label{sec:app_train}
We employ LLaMA-Factory~\citep{zheng2024llamafactory} for model fine-tuning. All models, including our reproduced baselines, are fine-tuned over 1 epochs with a batch size of 256 using 8 NVIDIA A100 GPUs. The peak learning rate is set to 5e-6, featuring a linear warm-up during the initial 3\% of training steps, followed by a cosine decay. The maximum sequence length is configured to 4096.

\section{Evaluation}
\label{sec:app_eval}
We compare GRA with baselines on the following 10 benchmarks:
\begin{itemize}
    \item The \textbf{GSM8K} dataset~\citep{cobbe2021gsm8k} contains 8,500 grade school-level math word problems.focus on real-world scenarios, testing a model's ability to parse natural language, extract relevant information, and perform accurate calculations.
    \item The \textbf{MATH} dataset~\citep{hendrycks2021math} evaluates mathematical reasoning with problems spanning arithmetic, algebra, calculus, and geometry, and 5 levels of difficulty.
    \item The \textbf{HumanEval} dataset~\citep{chen2021evaluating} is designed to evaluate the functional correctness of code generation models. It consists of 164 hand-written programming problems, each requiring the generation of a Python function to solve a specific task. The dataset assesses the ability of models to produce syntactically correct and logically accurate code.
    \item The \textbf{MBPP} dataset~\citep{austin2021program} contains around 1,000 Python programming tasks, focusing on fundamental programming concepts. It is split into training and test sets, with tasks ranging from simple operations to more complex algorithmic challenges, making it suitable for evaluating code generation and understanding.
    \item The \textbf{Hellaswag} dataset~\citep{zellers2019hellaswag} is a commonsense reasoning benchmark that includes 70,000 multiple-choice questions derived from everyday scenarios. It tests a model's ability to predict the most plausible continuation of a given context, emphasizing natural language understanding and reasoning.
    \item The \textbf{ARC-C} dataset~\citep{clark2018think} is a collection of science exam questions designed to evaluate advanced reasoning capabilities. It includes 2,590 questions across various science topics, requiring multi-step reasoning and knowledge application to solve.
    \item The \textbf{GPQA} datase~\citep{rein2023gpqa}t is a high-quality benchmark for evaluating general knowledge and reasoning. It consists of diverse questions spanning multiple domains, including science, history, and culture, designed to test the breadth and depth of a model's understanding.
    \item The \textbf{IFEval} dataset~\citep{zhou2023instruction} focuses on evaluating instruction-following capabilities in language models. It includes a variety of tasks that require models to follow complex instructions accurately, testing their ability to understand and execute multi-step directives.
    \item The \textbf{MMLU} dataset~\citep{hendrycks2020measuring} is a comprehensive benchmark covering 57 tasks across diverse domains, including humanities, STEM, and social sciences. It evaluates a model's ability to perform well across a wide range of subjects and tasks.
    \item The \textbf{BBH} dataset~\citep{suzgun2022challenging} is a subset of the Big-Bench benchmark, consisting of 23 challenging tasks that are particularly difficult for language models. These tasks require advanced reasoning, knowledge, and problem-solving skills, making them a robust test of model capabilities.
    
In our testing, we employ the default metrics of OpenCompass~\citep{2023opencompass} corresponding to each benchmark.
\end{itemize}

\section{Cases}
\setcounter{table}{1}
\renewcommand{\tablename}{Case}

A comprehensive data synthesis case is detailed in Case~\ref{case:GRA_Data_Synthesis_Case}.
\label{sec:Cases}
\begin{table*}
\begin{tcolorbox}[colback=gray!5,colframe=black!75, title=GRA Data Synthesis Cases]
\small
\# Sample Data: \\
\#\# Instruction1: \\
You are \textcolor{OliveGreen}{Cleopatra}, the last active ruler of the \textcolor{OliveGreen}{Ptolemaic} Kingdom of \textcolor{OliveGreen}{Egypt}. Reflect on your complex relationships with Julius Caesar and Mark Antony, and explain how these alliances influenced your rule and the fate of Egypt. Discuss the political and emotional aspects of these relationships, and what you would consider to be the most significant outcomes.\\
\#\# Response1: \\
As \textcolor{OliveGreen}{Cleopatra}, the last Pharaoh of ancient \textcolor{OliveGreen}{Egypt}, my life was a tapestry woven with threads of power, passion, and ambition. My relationships with Julius Caesar and Mark Antony were not only pivotal to my personal story but also crucial to the fate of my kingdom...\\
\#\# Instruction2: \\
Please use the identity of an old Guangzhou resident and use a \textcolor{OliveGreen}{humorous} and exaggerated tone to talk about your views on the changes in Guangzhou cuisine. In your answer, please use at least three \textcolor{OliveGreen}{metaphors} or \textcolor{OliveGreen}{personifications}, and make sure that each sentence is no less than 20 words, and must be separated by colons or semicolons.\\
\#\# Response2: \\
Oh, talking about the changes in Guangzhou's food, it's like a magic trick in the Xiguan Mansion, with surprises hidden behind every door; the Diandude on Shamian Island was like a happy fruit for the neighbors. Although there are some young competitors now, it is still going strong and exudes a unique charm...\\\\
\# Generate Key words: \\
\textcolor{OliveGreen}{Fantasy \quad Adventure \quad Quest}\\

\# Reference Summary:\\
1. Cleopatra's alliances with Caesar and Antony shaped Egypt's fate.\\
2. Humorous Guangzhou local describes food evolution\\\\
\# Generate Data:\\
\#\#Instruction:\\
Embark on a quest through the mystical realm of Eldoria, where you must navigate through enchanted forests, solve ancient riddles, and face fearsome dragons to retrieve the legendary Crystal of Aether. In this role-play, you are an aspiring hero, a seeker of truth, and a guardian of balance. Your adventure begins in the quaint village of Everlight, where whispers of the Crystal's whereabouts have begun to stir. What will you do first on your journey?\\
\#\#Response:\\
As an aspiring hero on this quest to find the legendary Crystal of Aether, I would first consult the village elders of Everlight. They are likely to have valuable insights or clues about the whereabouts of the Crystal and the nature of the quest. My approach would be to: 1. Seek out the village elders and listen to their stories and advice. 2. Offer to help with any local tasks or problems as a gesture of goodwill and to build relationships.3. Inquire about the ancient riddles and the lore surrounding the Crystal of Aether. 4. Ask if there are any artifacts or maps in the village that could assist on the journey. 5. Formulate a plan of action based on the information gathered, considering the potential routes through the enchanted forests and the locations of the fearsome dragons. This initial step is crucial for gathering intelligence and preparing for the challenges ahead.\\\\
\#Data Review:\\
\#\#Evaluation:\\
The average score of \textcolor{OliveGreen}{8.2} meets the criteria, but the standard deviation of \textcolor{red}{1.6} does not, therefore it is sent to the Data Adjudicator for evaluation.\\\\
\#Data Adjudicator:\\ 
\#\#Comment:\\
1. Clear plan, relevant steps, and coherent structure. Minor redundancy noted.\\
2. Response offers a thorough and strategic approach to consulting the village elders, ensuring a well-informed and well-prepared adventure.\\
3. The response provides a clear and structured plan to gather information, build relationships, and prepare for the journey. However, some minor verbalizations are present in the points, but are well approached despite a bit of excess wording in some areas.\\
\#\#Score:\\
The average score of \textcolor{OliveGreen}{8.1} meets the criteria.\\\\
\#Generate Data Summary:\\
Navigate enchanted forests, solve riddles, face dragons to retrieve Crystal of Aether.
\end{tcolorbox}
\caption{A comprehensive data synthesis case of GRA.}
\label{case:GRA_Data_Synthesis_Case}
\end{table*}

\section{Prompts}
\label{sec:prompt}
\setcounter{table}{0}
\renewcommand{\tablename}{Prompt}
The prompts we used, including domain identification, summary generation, key word extraction, key word generation, instruction generation, instruction evaluation,response generation, response evaluation and data adjudication are shown in Prompt~\ref{Domain Identification},~\ref{summary Generation},~\ref{Key Word Extraction},~\ref{Key Word Generation},~\ref{Instruction Evaluation},~\ref{Response Generation},~\ref{Response Evaluation},~\ref{Data Adjudication}.

\begin{table*}[htbp]
\begin{tcolorbox}[colback=gray!5,colframe=black!75, title=Domain Identification]
\small
\#\# role\\
- You are an expert in identifying the domain of SFT instructions.\\
\\
\#\# goal\\
- Determine the domain of the provided instructions using the specified classification criteria.\\
\\
\#\# classification criteria\\
- **Math**: Involves performing mathematical operations, applying formulas, and solving problems.\\
- **Coding**: Encompasses understanding, generating, debugging, and fixing code.\\
- **QA**: Provide specialized and accurate answers to questions within a specific domain or field..\\
- **Language**: Accurately understand and analyze the given text, and then perform tasks including translation, summarization, classification, etc.\\
- **Role Play**: Engage in simulated conversations or behaviors by assuming different roles to explore various perspectives or scenarios.\\
- **Reasoning**: Carry out complex causal reasoning and logical judgment, and solve multi-level complex problems through reasonable inference and analysis.\\
- **Creation**: Emphasize originality and creativity, create content according to requirements, and flexibly adapt to different styles and themes.\\
\\
\#\# output format\\
- Assign each data to one `domain` category mentioned above only.\\
- The output `domain` must be in English rather than Chinese.\\
- Use `<boa>` and `<eoa>` to encapsulate the domain, e.g., `<bod>{"domain":"the domain"}<eod>`.\\

\end{tcolorbox}
\caption{Prompt used in domain identification.}
\label{Domain Identification}
\end{table*}
\begin{table*}[htbp]
\begin{tcolorbox}[colback=gray!5,colframe=black!75, title=Summary Generation]
\small
\#\# role\\
- You are an expert in concise summarization.\\
\#\# goal\\
- Analyze the provided `instruction`, summarize and generate a brief, accurate `summary`.\\
\#\# requirement\\
- Limit the `summary` to no more than 30 words.\\
- Ensure clarity and precision in capturing the essence of the instruction.\\
- - Symbolic representations involving mathematical data must be in LaTeX format.\\
\#\# output\_format\\
- The output `summary` must be in English rather than Chinese.\\
- Use `<bod>` and `<eod>` to encapsulate the `summary`, e.g., `<bod>{"summary":"the summary"}<eod>`.\\
- Output only the generated `summary`, nothing else.\\
\end{tcolorbox}
\caption{Prompt used in summary generation.}
\label{summary Generation}
\end{table*}

\begin{table*}[htbp]
\begin{tcolorbox}[colback=gray!5,colframe=black!75, title=Key Word Extraction]
\small
\#\# role\\
- You are an expert at distilling key information from SFT instructions.\\
\#\# goal\\
- Extracts the `keywords` from the given instructions.\\
\#\# output\_format\\
- The number of `keywords` must not exceed three.\\
- The output `keywords` must be in English rather than Chinese.\\
- Directly and Only output the data `keywords`.\\
- Use `<bok>` and `<eok>` to encapsulate the `keywords`, e.g.,`<bok>{"keywords":["the keywords"]}<eok>`.\\
\end{tcolorbox}
\caption{Prompt used in key word extraction.}
\label{Key Word Extraction}
\end{table*}

\begin{table*}[htbp]
\begin{tcolorbox}[colback=gray!5,colframe=black!75, title=Key Word Generation]
\small
\#\# role\\
- You are a creative language model specializing in innovative `keyword` generation and `domain` synthesis.\\
\#\# goal\\
- Generate a unique set of `keywords` and `domain` that aligned with the provided examples and introducing novel concepts within the same domain.\\
\#\# requirement\\
- Maintain a single domain with no more than three keywords.\\
- Must only create one new data abstract.\\
- The keywords must be novel, different to the provided examples.\\
\#\# output\_format\\
- The output must be in English rather than Chinese.\\
- Use `<boa>` and `<eoa>` to encapsulate the domain and keywords, e.g., `<boa>{"domain":"the domain","keywords":["the keywords"]}<eoa>`.\\
- Directly output and only output the data abstract, nothing else.\\
\end{tcolorbox}
\caption{Prompt used in key word generation.}
\label{Key Word Generation}
\end{table*}

\begin{table*}[htbp]
\begin{tcolorbox}[colback=gray!5,colframe=black!75, title=Instruction Generation]
\small
\#\# role\\
- You are an expert in data generation.\\
\#\# goal\\
- Generate a high-quality `{domain}` related prompt, which related to key words like {keywords}.\\
\#\# requirement\\
- The generated prompt must be high-quality, accurately reflecting the domain and keywords.\\
- The generated prompt must be semantically complete and contain the notation and terminology specifications of the corresponding field.\\
- The generated prompt must specify the task the user wants the model to complete, answer a question, or perform an action.\\
- Symbolic representations involving mathematical data must be in LaTeX format.\\
\#\# output\_format\\
- Please ensure that the prompt generated are not overly verbose while maintaining quality and compliance.\\
- Output directly and only the prompt you generate, nothing else.\\
- The `generated prompt` must be wrapped with <boi> and <eoi>, e.g., `<boi>this is the generated prompt<eoi>`.\\
<begin of reference>\\
{summaries}\\
<end of reference>\\
\end{tcolorbox}
\caption{Prompt used in instruction generation.}
\label{Instruction Generation}
\end{table*}

\begin{table*}[htbp]
\begin{tcolorbox}[colback=gray!5,colframe=black!75, title=Instruction Evaluation]
\small
\#\# role\\
You are a rigorous reviewer who is responsible for evaluating the quality of the `instruction` in an instruction-response pair.\\
\#\# goal\\
- For the given `instruction`, You need to evaluate it according to the several evaluation dimensions specified in the `rule`.\\
\#\# rule\\
- **Reasonable**: `instruction` must be clear and executable to ensure that they can be effectively responded to and handled.\\
- **Completeness**:  `instruction` must provide all necessary information and context required to perform the task, ensuring there are no missing elements that could hinder understanding or execution.\\
- **Clarity**: `instruction` must be clear in language, accurate in wording, concise in syntactic structure, avoid vague expressions, clearly convey expected results, and eliminate ambiguity.\\
\#\# constraint\\
- **Reasonable**: if the instruction meets the criteria, score 1; otherwise, score 0.\\
- **Completeness**: if the instruction meets the criteria, score 1; otherwise, score 0.\\
- **Clarity**: if the instruction meets the criteria, score 1; otherwise, score 0.\\
\#\# output\_format\\
1. Use `<bos>` and `<eos>`to encapsulate the scores for each category as the list in python, e.g., `<bos>[1,1,0]<eos>`.\\
2. Directly output and only output the scores, nothing else.\\
\end{tcolorbox}
\caption{Prompt used in instruction evaluation.}
\label{Instruction Evaluation}
\end{table*}

\begin{table*}[htbp]
\begin{tcolorbox}[colback=gray!5,colframe=black!75, title=Response Generation]
\small
\#\# role\\
You are a professional language model assistant that aims to strictly generate high-quality `response` based on the `instruction`.\\
\#\# goal\\
- You need to generate a correct, complete, and sufficient `response` based on the `instruction`, strictly following the the `instruction`.\\
\#\# output\_format\\
- Symbolic representations involving mathematical data must be in LaTeX format.\\
- Please output directly and only output `response`, nothing else.\\
\end{tcolorbox}
\caption{Prompt used in response generation.}
\label{Response Generation}
\end{table*}

\begin{table*}[htbp]
\begin{tcolorbox}[colback=gray!5,colframe=black!75, title=Response Evaluation]
\small
\#\# role\\
- You are a rigorous reviewer who is responsible for evaluating the quality of the Response in an Instruction-Response pair.\\
\#\# goal\\
- For the `given response`, You need to strictly score and evaluate each Response according to the several evaluation dimensions specified in the `rule`.\\
\#\# rule\\
- Correctness: The response should accurately fulfill the instruction's requirements with correct information and logic.\\
- Clarity: The response must be expressed in clear, understandable language, avoiding ambiguity.\\
- Completeness: The response should include all necessary details and explanations to fully address the instruction.\\
- Relevance: The response must stay focused on instruction and exclude irrelevant information.\\
- Coherence: The response should be logically structured and flow smoothly from one point to the next.\\
- Ethicality: The response must be safe, unbiased, and free from harmful or inappropriate content. \\
\#\# constraint\\
- For each evaluation dimension in `rule`, give an score from 1 to 10, and form a `score\_list`.\\
- In addition to the scores, you also need to provide specific `review` for the scores.\\
- The `review` must be brief, under 30 words and focusing on the key points. \\
\#\# output\_format\\
- Use `<bos>` and `<eos>` to encapsulate the `score\_list`, e.g., `<bos>[8,4,6]<eos>`.\\
- Use `<boc>` and `<eoc>` to encapsulate the `review`, e.g., `<boc>THE CONTENT OF review<eoc>`.\\
- Please output directly and only the `score\_list` and `review` in the specified output format, nothing else.\\
- **Note: Be particularly critical in your evaluation to ensure only responses of the highest quality receive high scores.**\\
\end{tcolorbox}
\caption{Prompt used in response evaluation.}
\label{Response Evaluation}
\end{table*}

\begin{table*}[htbp]
\begin{tcolorbox}[colback=gray!5,colframe=black!75, title=Data Adjudication]
\small
\#\# role\\
- You are a rigorous reviewer who is responsible for evaluating the quality of the Response in an Instruction-Response pair.\\
\#\# goal\\
- For the `given response`, you need to strictly score and evaluate each Response according to the evaluation dimensions specified in the `rule`, while integrating insights from the `prior evaluation`.\\
\#\# rule\\
- Correctness: The response should accurately fulfill the instruction's requirements with correct information and logic.\\
- Clarity: The response must be expressed in clear, understandable language, avoiding ambiguity.\\
- Completeness: The response should include all necessary details and explanations to fully address the instruction.\\
- Relevance: The response must stay focused on instruction and exclude irrelevant information.\\
- Coherence: The response should be logically structured and flow smoothly from one point to the next.\\
- Ethicality: The response must be safe, unbiased, and free from harmful or inappropriate content. \\
\#\# constraint\\
- Must analyze the requirements of `instruction`, the quality of `response`, and the `prior evaluation` in detail.\\
- For each evaluation dimension in `rule`, give an score from 1 to 10, and form a `score\_list`.\\
- In addition to the scores, you also need to provide specific `review` for the scores.\\
- The `review` must be brief, under 30 words and focusing on the key points. \\
\#\# output\_format\\
- Use `<bos>` and `<eos>` to encapsulate the `score\_list`, e.g., `<bos>[8,4,6]<eos>`.\\
- Use `<boc>` and `<eoc>` to encapsulate the `review`, e.g., `<boc>THE CONTENT OF review<eoc>.`\\
- Please output directly and only the `score\_list` and `review` in the specified output format, nothing else.\\
- **Note: Be particularly critical in your evaluation to ensure only responses of the highest quality receive high scores.**\\
\#\# prior evaluation\\
{review}\\
<begin of Instruction-Response pair>\\
\#\# instruction\\
{instruction}\\
\#\# response\\
{response}\\
<end of Instruction-Response pair>\\
\end{tcolorbox}
\caption{Prompt used in data adjudication.}
\label{Data Adjudication}
\end{table*}

\setcounter{table}{2}
\renewcommand{\tablename}{Table}

\begin{table*}[!t]
\renewcommand{\arraystretch}{1.15}
\resizebox{\textwidth}{!}{
\begin{tabular}{ccccccccccccccccc}
\hline
     &      & \multicolumn{2}{c}{Math}                    &  & \multicolumn{2}{c}{Coding}                  &  & \multicolumn{4}{c}{Reasoning}                                                             &  & \multicolumn{2}{c}{General}                 &  &                      \\ \cline{3-4} \cline{6-7} \cline{9-12} \cline{14-15}
SIZE &      & GSM8K                & MATH                 &  & Humaneval            & MBPP                 &  & HellaSwag            & ARC-C                & GPQA                 & BBH                  &  & MMLU                 & IFEval               &  & AVG              \\ \hline
\multicolumn{17}{c}{LLama3.1 8B}                                                                                                                                                                                                                                                     \\ \hline
\rowcolor{gray!20}
\multicolumn{17}{c}{Condor}                                                                                                                                                                                                                                                          \\ \hline
10K  & GRA1 & 64.90                & 11.26                &  & 48.78                & 46.60                &  & 59.10                & 69.49                & 24.75                & 64.46                &  & 39.55                & 43.17                &  & 47.21                \\
20K  & GRA2 & 65.88                & {\ul \textbf{16.22}} &  & {\ul \textbf{51.83}} & 47.40                &  & 65.83                & 70.51                & {\ul \textbf{28.79}} & 63.28                &  & 44.89                & 48.40                &  & 50.31                \\
30K  & GRA3 & {\ul \textbf{67.10}} & 15.38                &  & {\ul \textbf{51.83}} & 48.40                &  & 61.82                & 69.15                & 25.25                & 63.13                &  & {\ul \textbf{45.60}} & 48.03                &  & 49.57                \\
40K  & GRA4 & 65.88                & 14.64                &  & {\ul \textbf{51.83}} & 47.40                &  & 66.07                & 72.20                & 27.27                & {\ul \textbf{64.75}} &  & 42.90                & {\ul \textbf{54.49}} &  & 50.75                \\
50K  & GRA5 & 65.58                & 15.82                &  & 50.61                & {\ul \textbf{48.80}} &  & {\ul \textbf{69.22}} & {\ul \textbf{72.54}} & 26.77                & 63.48                &  & 45.22                & 53.54                &  & {\ul \textbf{51.60}} \\ \hline
\rowcolor{gray!20}
\multicolumn{17}{c}{Alpaca}                                                                                                                                                                                                                                                          \\ \hline
10K  & GRA1 & 63.91                & 10.22                &  & 48.17                & 47.20                &  & 61.57                & 69.83                & {\ul \textbf{29.80}} & 63.80                &  & 35.15                & 46.08                &  & 47.58                \\
20K  & GRA2 & 65.66                & 14.86                &  & {\ul \textbf{53.66}} & 47.00                &  & 64.41                & 71.19                & 27.78                & 64.62                &  & 45.05                & 50.16                &  & 50.44                \\
30K  & GRA3 & 65.13                & 11.02                &  & {\ul \textbf{53.66}} & 48.00                &  & 62.11                & 70.17                & 26.77                & 62.91                &  & 45.35                & 53.41                &  & 49.86                \\
40K  & GRA4 & {\ul \textbf{66.41}} & 15.16                &  & 50.61                & {\ul \textbf{49.20}} &  & 65.32                & {\ul \textbf{73.56}} & 29.29                & {\ul \textbf{64.78}} &  & 45.63                & 55.31                &  & 51.53                \\
50K  & GRA5 & 65.58                & {\ul \textbf{17.2}}  &  & 52.44                & 48.60                &  & {\ul \textbf{66.30}} & 71.53                & {\ul \textbf{29.80}} & 64.50                &  & {\ul \textbf{48.58}} & {\ul \textbf{56.80}} &  & {\ul \textbf{52.14}} \\ \hline
\rowcolor{gray!20}
\multicolumn{17}{c}{WizardLM}                                                                                                                                                                                                                                                        \\ \hline
10K  & GRA1 & 63.68                & 10.08                &  & 49.39                & 47.80                &  & 61.67                & 68.47                & {\ul \textbf{29.80}} & 63.35                &  & 42.87                & 48.47                &  & 48.56                \\
20K  & GRA2 & 65.96                & 16.98                &  & 50.61                & 47.20                &  & 65.25                & 72.54                & 26.77                & {\ul \textbf{64.97}} &  & 44.07                & 51.44                &  & 50.58                \\
30K  & GRA3 & {\ul \textbf{68.16}} & 14.50                &  & {\ul \textbf{52.44}} & {\ul \textbf{50.00}} &  & 64.49                & 69.15                & 28.28                & 63.13                &  & {\ul \textbf{44.93}} & 52.28                &  & 50.74                \\
40K  & GRA4 & 65.88                & {\ul \textbf{18.98}} &  & 51.22                & 47.00                &  & {\ul \textbf{66.54}} & 72.88                & 26.77                & 63.28                &  & 43.10                & 53.52                &  & {\ul \textbf{50.92}} \\
50K  & GRA5 & 65.66                & 14.14                &  & 47.56                & 48.20                &  & 65.54                & {\ul \textbf{73.22}} & 28.28                & 64.25                &  & 42.14                & {\ul \textbf{55.27}} &  & 50.43                \\ \hline
\multicolumn{17}{c}{Qwen2.5 7B}                                                                                                                                                                                                                                                      \\ \hline
\rowcolor{gray!20}
\multicolumn{17}{c}{Condor}                                                                                                                                                                                                                                                          \\ \hline
10K  & GRA1 & 49.36                & 20.10                &  & 72.56                & 64.40                &  & 62.69                & 62.37                & 19.70                & 41.18                &  & 54.70                & 50.13                &  & 49.72                \\
20K  & GRA2 & 60.88                & 42.10                &  & 75.00                & 66.40                &  & 73.56                & 74.24                & {\ul \textbf{26.77}} & 51.95                &  & 63.70                & 50.97                &  & 58.56                \\
30K  & GRA3 & 65.88                & 34.30                &  & 78.05                & 66.00                &  & 77.74                & 73.90                & {\ul \textbf{26.77}} & 57.26                &  & 64.08                & 53.52                &  & 59.75                \\
40K  & GRA4 & {\ul \textbf{67.25}} & 39.04                &  & {\ul \textbf{79.27}} & {\ul \textbf{66.60}} &  & 74.23                & 73.22                & 26.26                & 51.11                &  & 63.12                & 52.86                &  & 59.30                \\
50K  & GRA5 & 64.75                & {\ul \textbf{42.82}} &  & 76.22                & 65.60                &  & {\ul \textbf{79.90}}  & {\ul \textbf{75.59}} & 26.26                & {\ul \textbf{61.49}} &  & {\ul \textbf{64.59}} & {\ul \textbf{54.02}} &  & {\ul \textbf{61.13}} \\ \hline
\rowcolor{gray!20}
\multicolumn{17}{c}{Alpaca}                                                                                                                                                                                                                                                          \\ \hline
10K  & GRA1 & 52.08                & 20.42                &  & 71.34                & 65.60                &  & 70.68                & 70.17                & 19.19                & 45.00                &  & 57.43                & 50.37                &  & 52.23                \\
20K  & GRA2 & 63.68                & 33.50                &  & 73.17                & {\ul \textbf{65.80}} &  & 71.59                & 71.19                & {\ul \textbf{24.24}} & 45.49                &  & 58.66                & 49.91                &  & 55.72                \\
30K  & GRA3 & 70.28                & 33.82                &  & {\ul \textbf{76.83}} & 64.80                &  & 80.23                & 74.24                & {\ul \textbf{24.24}} & 60.57                &  & 65.42                & 53.16                &  & 60.36                \\
40K  & GRA4 & {\ul \textbf{72.78}} & {\ul \textbf{46.48}} &  & 75.00                & {\ul \textbf{65.80}} &  & 76.99                & 71.19                & {\ul \textbf{24.24}} & 45.27                &  & 61.86                & {\ul \textbf{55.53}} &  & 59.52                \\
50K  & GRA5 & 51.71                & 44.62                &  & 74.39                & 64.60                &  & {\ul \textbf{81.62}} & {\ul \textbf{77.63}} & 20.71                & {\ul \textbf{62.97}} &  & {\ul \textbf{65.06}} & 54.58                &  & {\ul \textbf{59.79}} \\ \hline
\rowcolor{gray!20}
\multicolumn{17}{c}{WizardLM}                                                                                                                                                                                                                                                        \\ \hline
10K  & GRA1 & 53.98                & 23.20                &  & 73.17                & 65.20                &  & 73.66                & 70.51                & 22.73                & 48.30                &  & 59.42                & 51.08                &  & 54.13                \\
20K  & GRA2 & 59.74                & 31.02                &  & 73.17                & {\ul \textbf{65.80}} &  & 76.44                & 73.22                & 20.71                & 53.02                &  & 63.12                & 50.59                &  & 56.69                \\
30K  & GRA3 & 66.72                & 27.70                &  & 75.61                & 64.80                &  & 78.47                & 74.24                & 23.74                & 53.38                &  & 64.60                & 51.17                &  & 58.04                \\
40K  & GRA4 & {\ul \textbf{70.89}} & 38.58                &  & 76.22                & 64.60                &  & 79.02                & 77.63                & 19.19                & 59.29                &  & 65.10                & 53.55                &  & 60.41                \\
50K  & GRA5 & 58.98                & {\ul \textbf{47.84}} &  & {\ul \textbf{76.83}} & 64.60                &  & {\ul \textbf{81.58}} & {\ul \textbf{79.66}} & {\ul \textbf{28.79}} & {\ul \textbf{63.99}} &  & {\ul \textbf{65.23}} & {\ul \textbf{54.15}} &  & {\ul \textbf{62.17}} \\ \hline
\end{tabular}}
\caption{Detail Results of effect of the data iteration . Bold underlined values  represent the best results.}
\end{table*}

\end{document}